\documentclass[]{article}
\usepackage{authblk}

\usepackage[utf8]{inputenc} 
\usepackage[T1]{fontenc}    
\usepackage{url}            
\usepackage{booktabs}       
\usepackage{amsfonts}       
\usepackage{nicefrac}       
\usepackage{microtype}      
\usepackage{fullpage}

\usepackage{url}
\usepackage{graphicx}
\usepackage{amsmath}
\usepackage{bbm}
\usepackage{paralist}
\usepackage{Definitions}
\usepackage[algo2e,vlined,linesnumbered,algoruled]{algorithm2e}
\usepackage{algorithm}
\usepackage{algorithmic}
\usepackage{color}
\usepackage{enumitem}
\usepackage{mathtools}
\usepackage{subfig}
\usepackage{graphicx}
\usepackage{tabularx}
\usepackage[dvipsnames]{xcolor}

\newcommand{\Hawkes}{\ensuremath{\textrm{Hawkes}}}

\newcommand{\xhdr}[1]{\vspace{0mm}\noindent{{\bf #1. }}}

\newcommand{\explain}[2]{\underset{\mathclap{\overset{\uparrow}{#2}}}{#1}}
\newcommand{\explainup}[2]{\overset{\mathclap{\underset{\downarrow}{#2}}}{#1}}

\newcommand{\qb}{\boldsymbol{\omega}}
\newcommand{\qq}{{\omega}}
\newcommand{\thetab}{\boldsymbol{\theta}}

\newcommand\blfootnote[1]{%
  \begingroup
  \renewcommand\thefootnote{}\footnote{#1}%
  \addtocounter{footnote}{-1}%
  \endgroup
}

\title{Modeling the Dynamics of Online Learning Activity}

\author[1]{Charalampos Mavroforakis$^{*}$}
\author[2]{Isabel Valera}
\author[2]{Manuel Gomez-Rodriguez}

\affil[1]{Boston University, cmav@bu.edu}
\affil[2]{Max Planck Institute for Software Systems, ivalera@mpi-sws.org, manuelgr@mpi-sws.org}

\date{}

\begin{document}

\maketitle

\begin{abstract}
  People are increasingly relying on the Web and social media to find solutions to their problems in a wide range of 
domains.
In this online setting, closely related problems often lead to the same characteristic \emph{learning pattern}, in which people 
sharing these problems visit related pieces of information, perform almost identical queries or, more 
generally, take a series of similar \emph{actions}. 
In this paper, we introduce a novel modeling framework for clustering \emph{continuous-time} grouped streaming data, the 
hierarchical Dirichlet Hawkes process (HDHP), which allows us to automatically uncover a wide variety of learning patterns 
from detailed traces of learning activity.
%
Our model allows for efficient inference, scaling to millions of actions 
taken by thousands of users. 
Experiments on real data gathered from Stack Overflow reveal that our framework can recover meaningful learning patterns in terms 
of both content and temporal dynamics, as well as accurately track users'{} interests and goals over time.

\end{abstract}

\blfootnote{$^{*}$\scriptsize This work was done during the author'{}s internship at Max Planck Institute
for Software Systems.}

\section{Introduction}
%
%
Learning has become an online activity -- people routinely use a wide variety of online \emph{learning platforms}, ranging from wikis and question answering (Q\&A) sites to online 
communities and blogs, to learn about a large range of topics. 
In this context, people find solutions to their problems by looking for closely related pieces of information, 
executing a sequence of queries or, more generally, performing a series of online \emph{actions}.
%
%
%
%
For example, 
a high school student may study several closely related wiki pages to prepare an essay about a historical event;
a software developer may read several answers within a Q\&A site to solve a specific programming problem;
and, a researcher may check a specialized blog written by one of her peers to learn about a new concept or technique.
All the above are examples of \emph{learning patterns}, in which people perform a series of online actions -- reading 
a wiki page, an answer, or a blog -- to achieve a predefined goal -- writing an essay, solving a 
programming problem, or learning about a new concept or technique.
In this context, one may expect that people with similar goals undertake similar sequences of online actions and thus adopt similar
learning patterns.
Therefore, one could leverage the vast availability of online traces of users'{} learning activity to disambiguate among interleaved 
learning patterns adopted by individuals over time, as well as to automatically identify and track those people's interests and goals over time.
%
%

In this work, we introduce a novel probabilistic model, the Hierarchical Dirichlet Hawkes Process (HDHP), for clustering \emph{continuous-time} grouped streaming data, which we use to uncover the dynamics of  
learning activity on the web.
The HDHP leverages the properties of the Hierarchical Dirichlet Process (HDP)~\cite{Teh2006}, a popular Bayesian nonparametric model for 
clustering pro\-blems involving multiple groups of data, combined with the Hawkes process~\cite{hawkes1971spectra}, a
temporal point process particularly well fitted to model social activity~\cite{farajtabar2014activity,competing15icdm,zhou2013learning}.
%
In particular, the former is used to account for an infinite number of learning patterns, which are shared across users (groups) of an online learning platform. 
%
%
The latter is used to capture the temporal dynamics of the learning activity, which alternate between bursts of rapidly occurring events and 
relatively long periods of inactivity~\cite{barabasi05human}.

In more detail, in the case of the HDHP, the learning pattern distribution that determines the content and temporal parameters 
of each learning pattern is drawn from a Dirichlet process (DP)~\cite{ferguson1973bayesian}. 
Each user{}'s learning activity is modeled as a multivariate Hawkes process, with as many dimensions as the number of learning patterns (\ie, infinite), whose parameters are given by the DP and thus shared across all the users.   
Every time a user decides to perform a new action, she may opt for starting a new \emph{task} or follow-up on one of her previous ones. Here, 
tasks refer to sequences of related actions performed closely in time, which in turn can be viewed as realizations of learning patterns. 
%
%
Our model allows for an efficient inference algorithm, based on sequential Monte Carlo~\cite{smith2013sequential}, which scales to millions of 
actions and thousands of users.
%
%
We experiment on real-world data from Stack Overflow, using $\sim$$1.6$ million questions performed by
 $16{,}000$ users over a four year period. Our results show that our model, taking advantage of the temporal information in the data, does not only allow us to accurately track users{}' interest over time, but also provides more meaningful learning patterns in terms of content (20\% gain in perplexity) compared to the HDP.
A Python implementation of the proposed HDHP is available in GitHub.\footnote{\url{https://github.com/Networks-Learning/hdhp.py}}

\xhdr{Related work} 
The Hierarchical Dirichlet Hawkes process (HDHP) can be viewed as a model for clustering grouped {continuous-time} streaming data. In our application domain, each group of data corresponds 
to a user'{}s online actions and the clusters correspond to learning patterns, shared across all the users. 
Therefore, our work relates, on the one hand, to mo\-dels for clustering groups of data~\cite{LDA, HDP}, and, on the other hand, to models for clustering (single) streaming 
data~\cite{CRCP, DDCRP, ahmed11,DHP_Du2015}. 
To the best of our knowledge, models for clustering grouped streaming data are nonexistent to date.

The most popular models for clustering groups of data, the Latent Dirichlet Allocation (LDA)~\cite{LDA} and, its nonparametric counterpart, the HDP~\cite{Teh2006}, originate from the topic modeling literature. There, each document is transformed into a \textit{bag-of-words} and is modeled as a mixture of topics that are shared across all documents. 
More generally, models for clustering groups of data typically consider that each observation (word) within a group (document) is a sample from a mixture model, and the mixture components are shared 
among groups. 
One could use such models to cluster users'{} activity on the Web, however observations are assumed to be exchangeable and thus these models cannot account for the temporal dynamics of learning activity.
As a consequence, they are unable to track users'{} interests and goals over time.
%

Models for clustering streaming data can incorporate temporal dynamics~\cite{ahmed11, CRCP}, however they can only handle a single stream of data 
and thus cannot be used to jointly model several users'{} learning activity.
Additionally, most of these models discretize the time dimension into bins~\cite{ahmed11, CRCP}, introducing additional tuning parameters, and ignoring the self-excitation across events~\cite{DDCRP}, 
a phenomenon regularly observed in social activity data~\cite{farajtabar2014activity}.
Perhaps the most closely related work to ours is the recently proposed Dirichlet Hawkes Process (DHP)~\cite{DHP_Du2015}, a continuous-time model for streaming data that allows for self-excitation. 
However, DHP suffers from a significant limitation: the lack of an underlying DP (or, in fact, any other Bayesian nonparametric) prior on the cluster distribution compromises the identifiability and reproducibility of the model.
Additionally, from the perspective of our application, DHP only allows for single data stream (user) and enforces clusters to be forgotten after some time. The latter is an overly restrictive assumption, since a user 
may perform similar actions, \ie, belonging to the same learning pattern, 
over widely spaced intervals of time. 
\section{Preliminaries}
%
%
In this section, we briefly review the major building blocks of the Hierarchical Dirichlet Hawkes process (HDHP): the Hierachical Dirichlet process (HDP)~\cite{Teh2006} and the Hawkes process~\cite{aalen2008survival}.
\subsection{Hierarchical Dirichlet Process}\label{sec:HDP}
The {HDP} is a Bayesian nonparametrics prior, useful for clustering grouped data~\cite{Teh2006}, which allows for an unbounded number of clusters whose parameters are shared across all the groups. 
It has been broadly applied for topic modeling as the nonparametric counterpart of the Latent Dirichlet Allocation (LDA), where the number of topics is finite and predefined. 
%
%
More specifically, this process defines a hierarchy of Dirichlet processes (DPs), in which a set of random probability measures $G_j \sim DP(\beta_1,G_0)$ (one for each group of data) are distributed as DPs with concentration parameter $\beta_1$ and base distribution $G_0$. The latter is also distributed as a DP, \ie, $G_0\sim DP(\beta_0,H)$. 
In the HDP, the distributions $G_j$ share the same support as $G_0$, and are conditionally independent given $G_0$. 


\xhdr{Chinese Restaurant Franchise}
An alternative representation of the HDP is the Chinese Restaurant Franchise Process (CRFP), which allows us not only to obtain samples from the HDP but also to derive efficient inference 
algorithms. 
%
%
The CRFP assumes a franchise with as many restaurants as the groups of data (\eg, number of documents), where all of the restaurants share the same menu with an infinite number of dishes (or clusters). 
%
%
In particular, one can obtain samples from the HDP as follows:
\vspace{1mm}
\begin{compactitem}
\item[1.] Initialize the total number of dishes $L=0$, and the total number of tables in each restaurant $K_r=0$, for $r=1,\ldots, R$, with $R$ being the total number of restaurants in the franchise.
\item[2.] For each restaurant $r=1, \ldots, R$:
\item[] For customer $i=1, \ldots, N_r$ ($N_r$ is the total number of customers entering restaurant $r$):
\begin{compactitem}[--]
	\item Sample the table for the $i$-th customer in restaurant $r$ from a multinomial distribution with probabilities
	\begin{equation}
	\begin{array}{llll}
	\Pr(b_{ri} = k) & = & \frac{n_{rk}}{\beta_1+i-1} & \text{for } \ k=1,\ldots,K_r \\ 
	\Pr(b_{ri} = K_r +1) & = & \frac{\beta_1}{\beta_1+i-1} & 	
	\end{array}
	\vspace{1mm}
	\end{equation}	
	 where $n_{rk}= \sum_{j=1}^{i-1} \mathbbm{I}( b_{rj} =k)$ is the number of customers seated at the $k$-th table of the $r$-th restaurant. 
	 
	 \item  If $b_{ri}=K_r+1$, \ie, the $i$-th customer sits at a new table, sample its dish from a multinomial distribution with probabilities 
	 \begin{equation}
	 \begin{array}{llll}
	 \Pr(\phi_{r(K_r+1)} = \varphi_{\ell}) & =  & \frac{m_\ell}{K + \beta_0} & \text{for } \ell=1,\ldots,L \\ 
	 \Pr(\phi_{r(K_r+1)} = \varphi_{L+1}) & = & \frac{\beta_0}{K + \beta_0} & 
	 \end{array}
	 \vspace{1mm}
	 \end{equation}
	 %
where $K=\sum_{j=1}^{r} K_j $ is the total number of tables in the franchise,  $m_\ell= \sum_{j=1}^{r} \sum_{k=1}^{K_j} \mathbbm{I}(\phi_{jk} = \varphi_\ell)$ is the total number of tables serving dish $  \varphi_\ell$ in the franchise, and $\varphi_{L+1}\sim H(\varphi)$ is the new dish, \ie,  the parameters of the new cluster. 
	
	\item Increase the number of tables in the $r$-th restaurant $K_r=K_r+1$ and, if  $\phi_{rK_r}=\varphi_{L+1}$ (\ie, the new table is assigned to a new dish/cluster), increase also the total number of clusters in the franchise $L=L+1$.
	%
	  \end{compactitem}
\end{compactitem}
Note that, although in the process above we have generated the data (customers) for each group (restaurant) sequentially, due to the exchangeability properties of the HDP, the resulting distribution of the data 
is invariant to the order at which customers are assumed to enter any of the restaurants~\cite{Teh2006}. 

\subsection{Hawkes Process}\label{sec:HP}
%
%
A Hawkes process is a stochastic process in the family of temporal point processes \cite{aalen2008survival}, whose realizations consist of lists of discrete events localized in time, 
$\cbr{t_1, t_2, \ldots, t_n}$ with $t_i \in \RR^+$. 
%
%
A temporal point process can be equivalently re\-pre\-sen\-ted as a counting process, $N(t)$, which records the number of events before time $t$. The probability of an event happening in a small time window $[t,t+dt)$ is given by $\mathbbm{P} (dN (t)= 1 |\Hcal(t) )=\lambda^*(t)dt$,
where $dN (t)\in \{0,1\}$ denotes the increment of the process, $\Hcal(t)$ denotes the history of events up to but not in\-clu\-ding time $t$, $\lambda^*(t)$ is the conditional intensity function (intensity, in short),
and the sign $^{*}$ indicates that the intensity may depend on the history $\Hcal(t)$.
In the case of Hawkes processes, the intensity function adopts the following form: 
\begin{equation}\label{eq:hawkes}
\lambda^*(t) = \mu + \sum_{t_i\in \Hcal(t)} \kappa_{\alpha}(t,t_i), 
\end{equation}
where $\mu$ is the base intensity and $\kappa_{\alpha}(t,t_i)$ is the triggering kernel, which is parametrized by $\alpha$. 
Note that this intensity captures the self-excitation phenomenon across events and thus allows modeling bursts of activity. As a consequence, the Hawkes process has been increasingly used to model social 
activity~\cite{farajtabar2014activity,competing15icdm,zhou2013learning}, which is characterized by bursts of rapidly occurring events separated by long periods of inactivity~\cite{barabasi05human}.
Finally, given the history of events  in an observation window $[0, T)$, denoted by $\Hcal(T)$,  we can express the log-likelihood of the observed data as 
\begin{align}
  \label{eq:loglikehood_fun}
  \Lfra_{T} = \sum_{i: t_i \in \Hcal(T)} \log \lambda^*(t_i) - \int_{0}^T \lambda^*(\tau)\, d\tau. 
\end{align}
\section{Learning activity model} \label{sec:model}
%
%
In an online learning platform, users find solutions to their problems by sequentially looking for closely related pieces of information within the site, executing a
sequence of queries or, more generally, performing a series of online \emph{actions}.
In this context, one may expect people with similar goals to undertake similar sequences of actions, which in turn can be viewed as realizations of an unbounded number of
\emph{learning patterns}.
Here, we assume that each action is linked to some particular content and we propose a modeling framework that characterizes sequences of actions by means
of the timestamps as well as the associated content of these actions.
%
Next, we formulate our model for online learning activity, starting by describing the data it is designed for.

\xhdr{Learning activity data and generative process} Given an online learning platform with a set of users $\Ucal$, we represent the learning actions of each user as a triplet 
%
%
\begin{equation}
e:=(~~\explainup{t}{\text{time}}~~,~~\explain{\qb}{\text{content}}~~, ~~\explainup{{\color{gray}p}}{\text{learning pattern} }~~), 
\end{equation}
which means that at time $t$ the user took an action linked to content $\qb$ and this action is associated to the learning pattern $p$, which is {\color{gray}hidden}.
%
%
%
Then, we denote the history of learning actions taken by each user $u$ up to, but not including, time $t$ as $\Hcal_u(t)$. 

%
We represent the times of each user $u$'{}s learning actions within the platform as a set of counting processes, $\mathbf{N}_u(t)$, in which the
$\ell$-th entry counts the number of times up to time $t$ that user $u$ took an action associated to the learning pattern $\ell$. Then, we characterize
these counting processes using their corresponding intensities as $\EE[d\mathbf{N}_u(t) | \Hcal_u(t)] = \lambdab^{*}_u(t)\,dt$,
%
%
where $d\mathbf{N}_u(t) = [dN_{u,\ell}(t)]_{\ell \in [L]}$ denotes the number of learning actions in the time window $[t, t+dt)$ for each learning pattern, $\lambdab^{*}_u(t) = [\lambda^*_{u,\ell}(t)]_{\ell \in [L]}$
denotes the corresponding pattern intensities, $L$ is the number of learning \-patterns, and the sign $^{*}$ indicates that the intensities may depend on the user{}'s history, $\Hcal_u(t)$.
%
%
Additionally, for each learning action $e = (t, \qb, p)$, the content $\qb$ is sampled from a distribution $f(\qb | p)$, which depends on the corresponding learning pattern $p$.
%
%
Here, in order to account for an unbounded number of learning patterns, \ie, $L \to \infty$, we assume\- that the
learning pattern distribution follows a Dirichlet process (DP).
Next, we specify the functional form of the user intensity associated to each learning pattern and the content distribution, and we elaborate further on the learning pattern distribution.

\xhdr{Intensity of the user learning activity}
Every time user $u$ performs a learning action, she may opt to either start a new \textit{task}, defined as a sequence of learning actions similar in content and performed
closely in time (\ie, a realization of a learning pattern),
or to follow-up on an already on-going task. The multivariate Hawkes process~\cite{aalen2008survival}, described in Section~\ref{sec:HP}, presents itself as a natural choice to model this behavior. This way, 
each dimension $\ell$ corresponds to a learning pattern $\ell$ and its associated intensity is given by 
\begin{equation}\label{eq:pattern_intensity}
\lambda^*_{u,\ell}(t) = \underbrace{~{\mu_u}~\pi_\ell~}_{\text{new task}} +  \overbrace{\sum\limits_{j: t_j\in \Hcal_u (t), ~ p_j=\ell} \kappa_{\ell}(t, t_j)}^{\text{follow-up}}. 
\end{equation}
Here, the parameter $\mu_u\geq 0$ accounts for the rate at which user $u$ starts new tasks, $\pi_\ell \in [0,1]$ is the probability that a user adopts learning pattern $\ell$ (referred to as
\emph{learning pattern popularity} from now on), and $ \kappa_{\ell}(t,t')$ is a nonnegative kernel function that models the decaying influence of past events in the pattern's intensity.
Due to convenience in terms of both theoretical properties and model inference, we opt for an exponential kernel function in the form $\kappa_{\ell}(t, t')= \alpha_\ell \exp(-\nu(t-t'))$,
where $\alpha_\ell$ controls the self-excitation (or burstiness) of the Hawkes process 
and $\nu$ controls the decay.
Finally, note that we can compute user $u$'{}s average intensity at time $t$ analytically as~\cite{farajtabar2014activity}:
\begin{equation} \label{eq:lambdab}
\EE_{\Hcal_u(t)}[\lambdab_u^{*}(t)] =\Big[e^{(\Ab-\nu I)t}+\nu (\Ab-\nu I)^{-1} \left( e^{(\Ab-\nu \Ib)t} - \Ib \right)\Big]  \mub_u , \vspace{3mm}
\end{equation}
where $\Ab =diag([\alpha_1, \ldots, \alpha_L])$, $\mub_u=[\mu_u \pi_1, \ldots, \mu_u \pi_L]^{\top}$, $\Ib$ is the identity matrix, and
the expectation is taken over all possible histories $\Hcal_u(t)$.
We can also compute the expected number of actions performed by user $u$ until time $T$ as 
\begin{equation} \label{eq:N_u}
\EE_{\Hcal_u(T)}[\Nb_u(T)] =  \int_0^T \EE_{\Hcal_u(\tau)}[\lambdab_u^{*}(\tau)] \, d \tau. 
\end{equation}

\xhdr{Content distribution}
%
We gather the content associated to each learning action $e=(t,\qb, p)$ as a vector $\qb$, in which each element is a word sampled from a vocabulary $\Wcal$ as 
\begin{equation}
\qq_{j} \sim Multinomial (\thetab_{p}), 
\end{equation}
where $\thetab_{p}$ is a $|\Wcal|$-length vector indicating the probability of each word to appear in content from pattern $p$.

\xhdr{Learning pattern parameters}
The distribution of the learning patterns is sampled from a DP, $G_0 \sim DP(\beta, H)$, which can be alternatively written as 
\begin{equation}
G_0 = \sum_{\ell=1}^\infty \pi_\ell \delta_{\varphi_\ell},
\end{equation}
where $\pib=(\pi_\ell)_{\ell=1}^{L=\infty} \sim GEM(\beta)$ is sampled from a stick breaking process~\cite{sethuraman:stick} and $\varphi_\ell= \{\alpha_\ell, \thetab_\ell \} \sim H(\varphi)$.

\xhdr{Remarks} Overall, the proposed learning activity model, which we refer to as the Hierarchical Dirichlet Hawkes process (HDHP), is based on a 2-layer hierarchical design. The top layer is a Dirichlet process
that determines the learning pattern distribution, and the bottom layer corresponds to a collection of independent multivariate Hawkes processes, one per user, with as many dimensions
as the number of learning patterns, \ie, infinite.
In the HDHP, the popularity of each learning pattern, or equivalently the probability of assigning a new task to it, is constant over time and given by the distribution $G_0$. However, the probability distribution 
of the learning patterns for each specific user evolves continuously over time and directly depends on her instantaneous intensity. 
Finally, we remark that, due to the infinite dimensionality of the Hawkes process that captures the learning activity of each user, sampling or performing inference directly on this model is intractable. Fortunately, we can benefit from the properties of both the Hawkes and the DP,
and propose an alternative generative process that we can then utilize to efficiently obtain samples of the HDHP.

%
%
\subsection{Tractable model representation}\label{generative}
%
%
Similarly to the HDP, we can  generate samples from the proposed HDHP by following a generative process similar to the CRFP.
To this end, we leverage the properties of the Hawkes process and represent the learning actions of all the users in the learning platform as a multivariate Hawkes process, with as many dimensions as the users, from which we sample the user and the timestamp associated to the next learning action.
This action is then assigned to either an existing or a new task with a probability that depends on the history of that user up to, but not including, the time of the action.
When initiating a new task, the associated learning pattern is sampled from a distribution that accounts for the overall popularity of all the learning patterns. 
We finally sample the action content $\qb$ as we discussed previously.

In the process described above, each user can be viewed as a restaurant, each action as a customer, each task as a table, and each pattern as a dish, as in the original CRFP.
%
%
Hence, if we assume a set of users $\Ucal$ and vocabulary $\Wcal$ for the content, we can generate $N$ learning actions as follows:
\begin{compactitem}
\item[1.] Initialize the \emph{total number of tasks} $K=0$ and the \emph{total number of learning patterns} $L=0$.

\item[2.] For $n = 1,\ldots, N$:
 \begin{compactitem}

 \item[(a)] Sample the time $t_n$ and user  $u_n\in \Ucal $ for the new action, such that $t_n>t_{n-1}$, as in~\cite{competing15icdm}
	 \begin{equation} \label{eq:samplet_u}
	 (t_n, u_n) \sim {\Hawkes}\left(
	 \begin{array}{c}
	 \mu_1 + \sum\limits_{\substack{i=1}}^{n-1}\kappa_{b_i}(t_n, t_i)  \mathbbm{I}(u_i = 1)\\
	 \vdots\\
	 \mu_{\Ucal} + \sum\limits_{\substack{i=1 }}^{n-1}\kappa_{b_i} (t_n, t_i) \mathbbm{I}(u_i = \Ucal)
	 \end{array}
	 \right)
	 \end{equation}
\item[(b)] Sample the task $b_n$ for the new action from a multinomial distribution with probabilities
%
	\begin{equation}
	\label{eq:b_sample}
	\begin{array}{llll}
	\Pr(b_n = k) & = &  \frac{\lambda^*_{u_n, k}(t_n)}{\lambda^*_{u_n} (t_n)},
	& \text{for } \ k=1,\ldots,K \vspace{1mm}\\ 
	\Pr(b_n = K+1) & = & \frac{\mu_{u_n}}
	{\lambda^*_{u_n} (t_n)}
	& 
	\end{array}
	\end{equation}
	 where $\lambda^*_{u_n, k}(t_n)=\sum_{i=1}^{n-1} \kappa_{b_i}(t_n,t_i) \mathbbm{I}(u_i = u_n, b_i = k)$, and $\lambda^*_{u_n} (t_n) = \mu_{u_n} + \sum_{i=1}^{n-1} \kappa_{b_i}(t_n,t_i) \mathbbm{I}(u_i = u_n)$ 
	 is the total intensity of user $u_n$ at time $t_n$. 
	\item[(c)]  If $b_n = K + 1$, assign the new task to a learning pattern with probability
	 \begin{equation}
	 \label{eq:phi_sample}
	 	 \begin{array}{llll}
	 	  \Pr(\phi_{K+1} = {\varphi_{\ell}}) & = & \frac{m_\ell}{K + \beta}, & \text{for } \ell=1,\ldots,L \vspace{1mm}\\ 
	 	 \Pr(\phi_{K+1} = \varphi_{L+1}) & = & \frac{\beta}{K + \beta} & 
	 	 \end{array}\raisetag{3\baselineskip}
	 \end{equation}
	 where $m_\ell= \sum_{k=1}^K \mathbbm{I}(\phi_k = \varphi_\ell)$ is the number of tasks assigned to learning pattern $\ell$ across all users, and $\varphi_{L+1}= \{\alpha_{L+1},\thetab_{L+1} \} $ is
	 the set of parameters of the new learning pattern $L+1$, which 
	  we sample from $\alpha_{L+1} \sim Gamma(\tau_1,\tau_2)$ and $\thetab_{L+1} \sim Dirichlet(\eta_0)$.
	 %
	 Then, increase the number of tasks $K=K+1$ and, if  $\phi_{K+1}=\varphi_{L+1}$, increase also the number of clusters $L=L+1$.
	%
	  \item[(d)] Sample each word in the content $\qb_n$ from $\omega_{n,j}\sim Multinomial(\thetab_{b_n})$.
	 \end{compactitem}
\end{compactitem}

\xhdr{Remarks} 
Note that, in the process above, both users and learning patterns are exchangeable. However, contrary to the CRFP, the generated data consist of a sequence of discrete events localized in time, which therefore do not satisfy the exchangeability property.  
Moreover, the complexity of this generative process differs from the standard CRFP only in two steps. First, it needs to sample the event time and user from a Hawkes process as in Eq.~\ref{eq:samplet_u}, which can be done in linear time with respect to the number of users~\cite{farajtabar2014activity}. Second, while the CRFP only accounts for the number of customers at each table, the above process needs to evaluate the intensity associated with each table (see Eq.~\ref{eq:b_sample}), which can be updated in O(1) using the properties of the exponential function. 

We also want to stress that although the above generative process resembles the one for the Dirichlet Hawkes process (DHP)~\cite{DHP_Du2015}, they differ in two key 
factors. 
First, the DHP can only generate a single sequence of events, while the above process can generate an independent sequence for each user. 
Second, the DHP does not instantiate an explicit prior distribution on the clusters, which results in a lack of identifiability and reproducibility of the model. 
In other words, new events in the DHP are only allowed to join a new or a currently active cluster -- once a cluster ``dies'' (\ie, its intensity becomes negligible), no new event can be assigned to it anymore. 
As a result, two bursts of events that are similar in content and dynamics but widely separated in time will be assigned to different clusters, leading to multiple copies of the same cluster. 
%
In contrast, our generative process ensures the identifiability and reproducibility of the model by placing a DP prior on the cluster distribution, and using the CRFP to integrate out the learning pattern popularity.



\subsection{Inference}%
%
Given a collection of $N$ observed learning actions 
performed by a set of users $\Ucal$ during a time period $[0, T)$, our goal is to infer the learning patterns that these actions belong to. 
To efficiently sample from the posterior distribution, similarly to~\cite{DHP_Du2015}, we leverage the generative process described in Section~\ref{generative}. We derive a sequential Monte Carlo (SMC) algorithm that exploits the temporal dependencies in the observed
data to sequentially sample the latent variables \-associated with each learning action.
In particular, the posterior distribution $p(b_{1:N} |t_{1:N}, u_{1:N}, \qb_{1:N})$ is sequentially approxi\-ma\-ted with a set of $\Pcal$ particles, which are sampled from a proposal distribution $q(b_{1:N} |  t_{1:N},  u_{1:N}, \qb_{1:N})$. To infer the global parameters, $\mu_u$ and $\alpha_{\ell}$, we follow the literature in SMC devoted to the estimation of a static parameter~\cite{cappe2007overview, carvalho2010particle}, and sequentially update the former by maximum likelihood estimation and the latter by sampling from its posterior distribution. 
The inference algorithm, which is detailed in Appendix~\ref{app:inference}, has complexity $O(\Pcal  (\Ucal + L + K)+ \Pcal )$ per observed learning action, where $L$ and $K$ are respectively the number of learning patterns and the number of tasks uncovered up to this action. 
%



\begin{figure}[t]
	\centering
	\subfloat[Estimation of $\mu_u$]{\makebox[0.25\textwidth][c]{\includegraphics[width=0.22\textwidth]{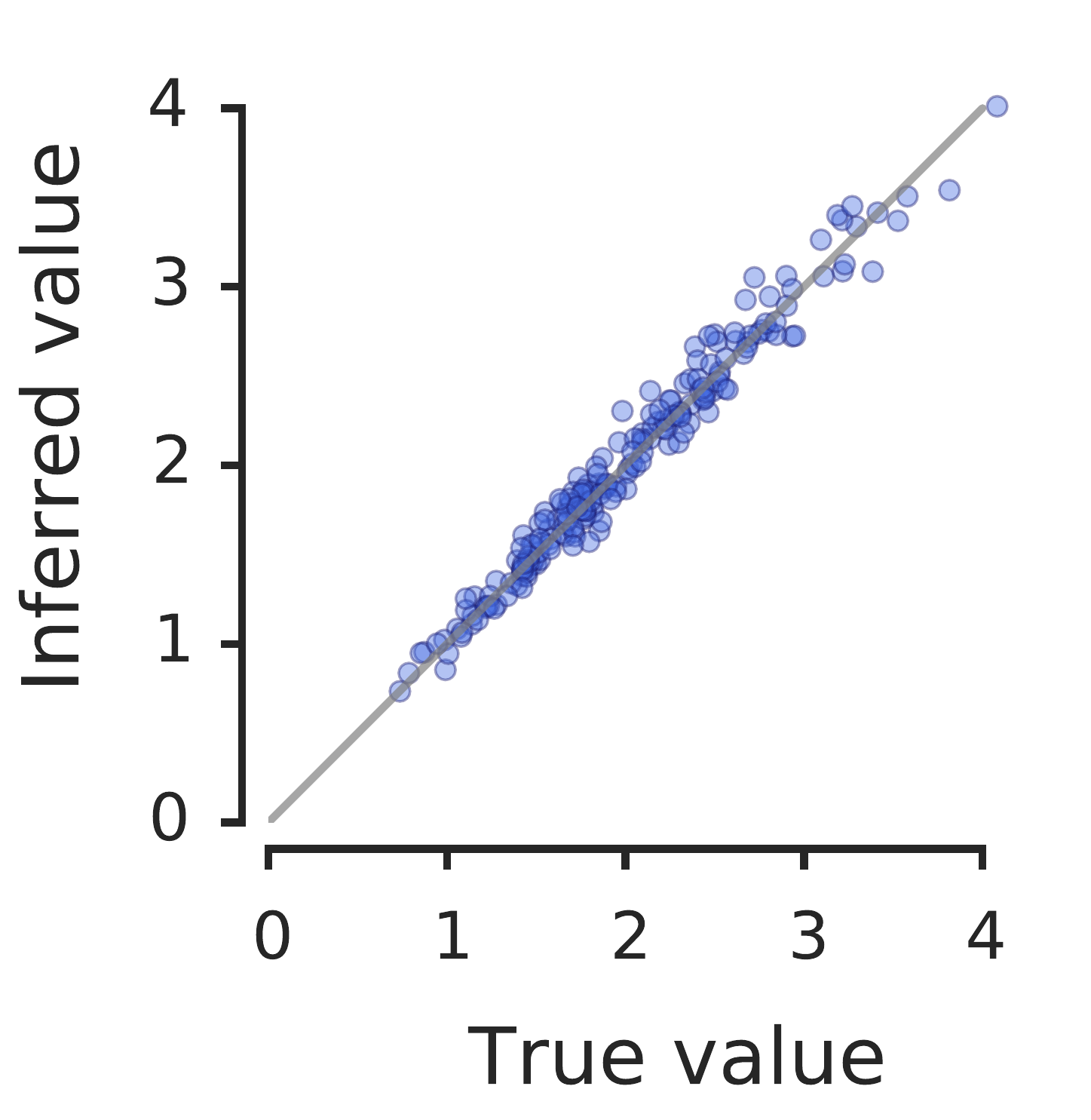}}\label{fig:estimation_mu}}
	\subfloat[Estimation of $\alpha_\ell$]{\makebox[0.25\textwidth][c]{\includegraphics[width=0.22\textwidth]{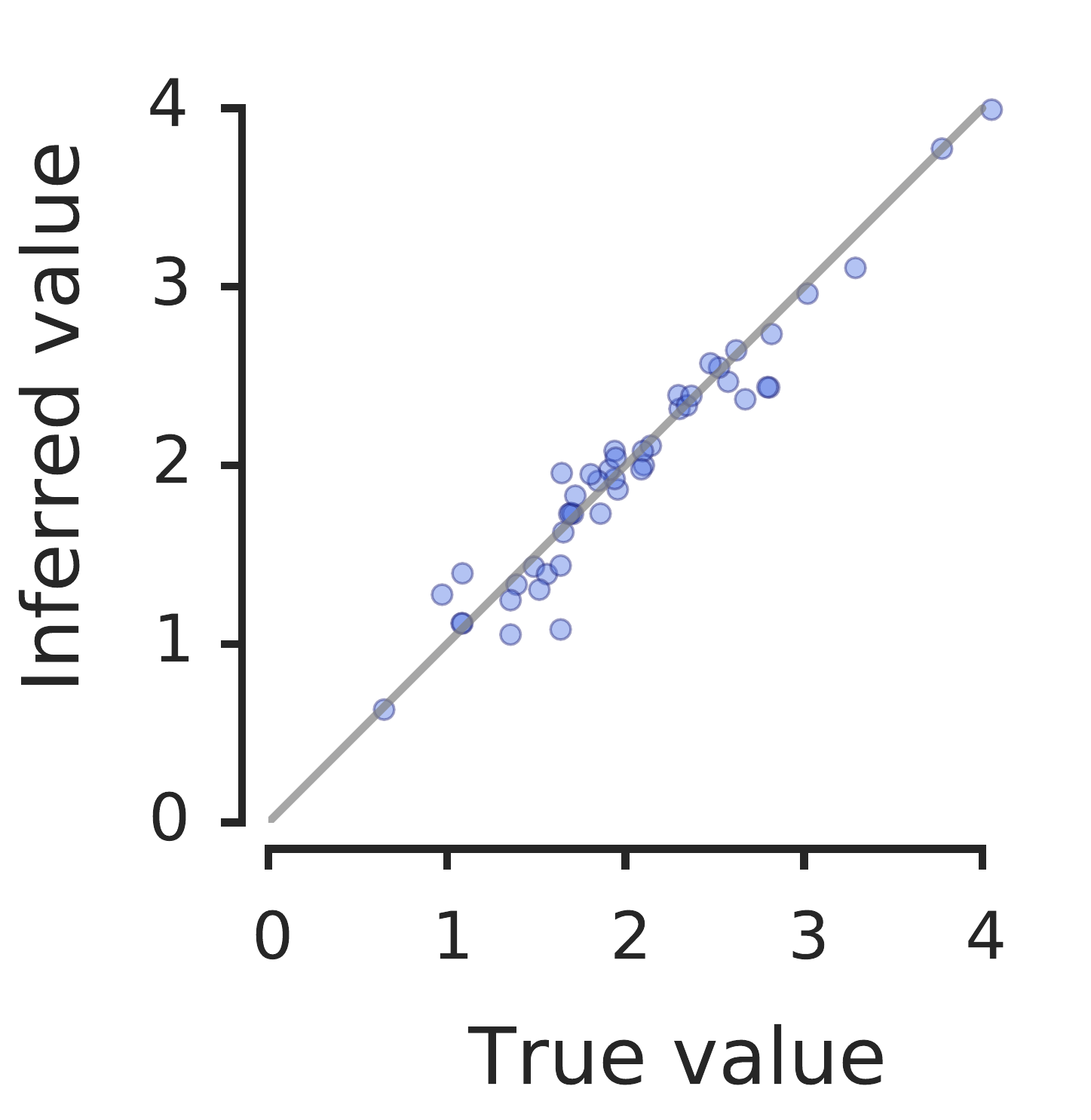}}\label{fig:estimation_alpha}}
	\subfloat[Clustering accuracy]{\makebox[0.22\textwidth][c]{\includegraphics[width=0.22\textwidth]{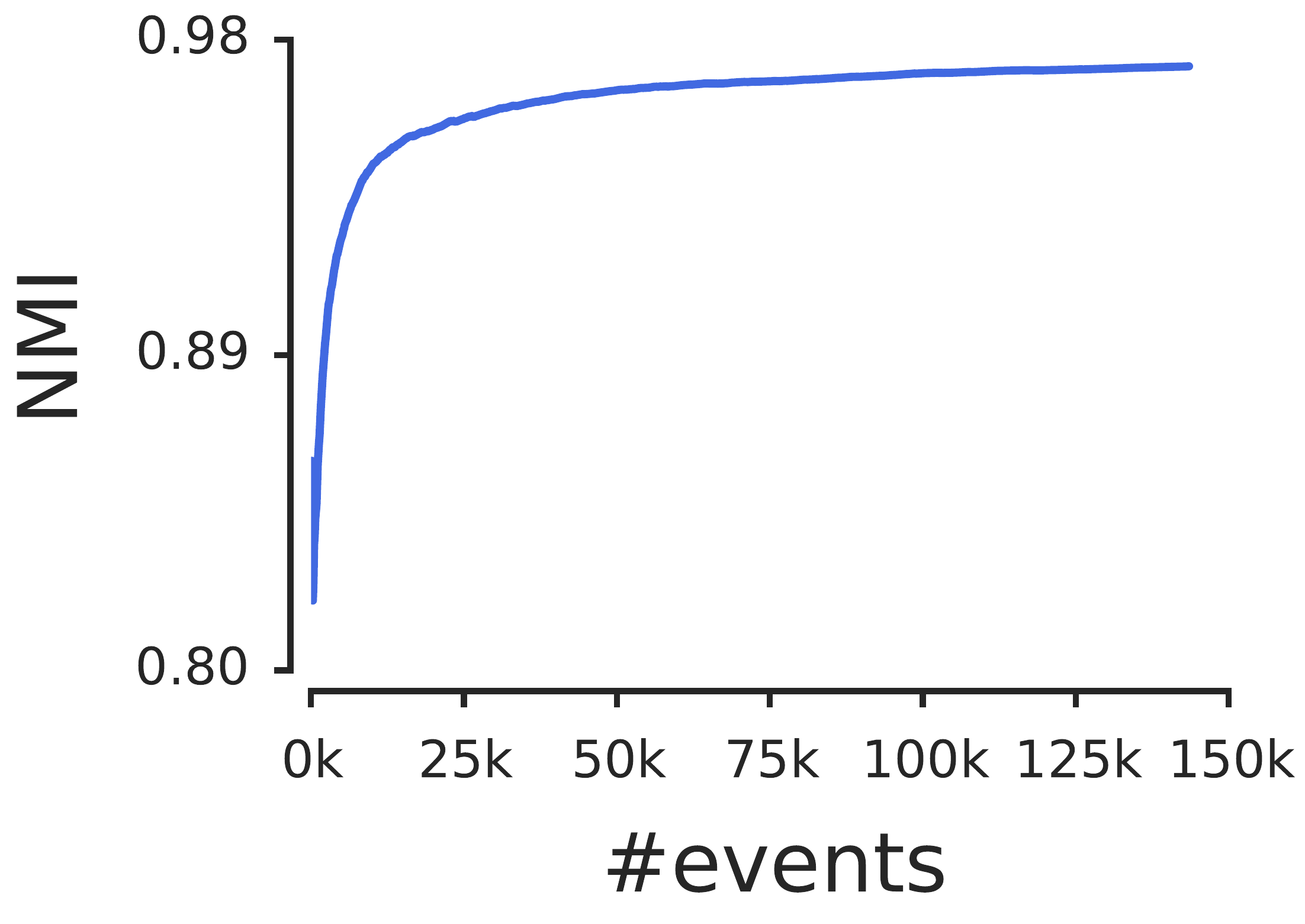}}\label{Clustering_quatlity}}
	%
	\caption{Evaluation of the inference algorithm at recovering the model parameters and latent learning pattern associated to each learning event on $150$k synthetically generated data. }\label{fig:synthetic}
	%
\end{figure}

\section{Experiments}

%
%

\subsection{Experiments on synthetic data}
%
In this section, we experiment with synthetic data and show that our inference algorithm can accurately recover the model parameters as well as assign each 
generated learning action to the true learning pattern given only the times and content of the learning events.

\xhdr{Experimental setup} 
We assume a set of $200$ users, $L= 50$  learning patterns and a vocabulary of size $|\Wcal| = 100$. We then sample the base intensity of each user $\mu_u$ from $Gamma(10, 0.2)$, and the learning pattern popularity vector $\pib$ from a Dirichlet distribution with concentration parameters equal to $1$. For each learning pattern, we sample the kernel parameter $\alpha_\ell$ from $Gamma(8, 0.25)$, we randomly pick $30$ words that will be used by the pattern and sample their distribution from a 
$Dirichlet$ distribution with parameters equal to $3$. We assume a kernel decay of $\nu= 5$. Then, for each user we generate online learning actions from the corresponding multivariate Hawkes process.

\xhdr{Results} Figure~\ref{fig:synthetic} summarizes the results by showing the true and the estimated values of the base intensity of each user $\mu_u$ and the kernel parameter of each pattern $\alpha_{\ell}$, using a total of $150$k events. 
Moreover, it also shows the normalized mutual information (NMI) between the true and inferred clusters of actions against the number of events seen by our inference algorithm. Here, we report the results for the particle which provided the maximum likelihood, and match the inferred learning patterns to the true ones by maximizing the NMI score. 
Our inference algorithm accurately recovers the model parameters and, as expected, using more events when inferring the model parameters leads to more accurate assignment of events to learning patterns.


%
%
%

\subsection{Experiments on real data}
\label{sec:real_experiments}
%
In this section, we experiment on real data gathered from Stack Overflow, a popular question answering (Q\&A) site, where users can post 
questions -- with topics ranging from C\# programming to Bayesian nonparametrics -- which are, in turn, answered by other users of the site. 
%
We infer our proposed HDHP on a large set of learning actions, and show 
that the proposed HDHP 
recovers meaningful learning patterns and allows us to accurately track users' interests over time.

%
%

\begin{figure}[t]
	\centering
		\subfloat[Dynamics]
	{\makebox[0.5\linewidth][c]{\includegraphics[width=0.35\linewidth, , trim=0 -1cm 2cm 2cm]{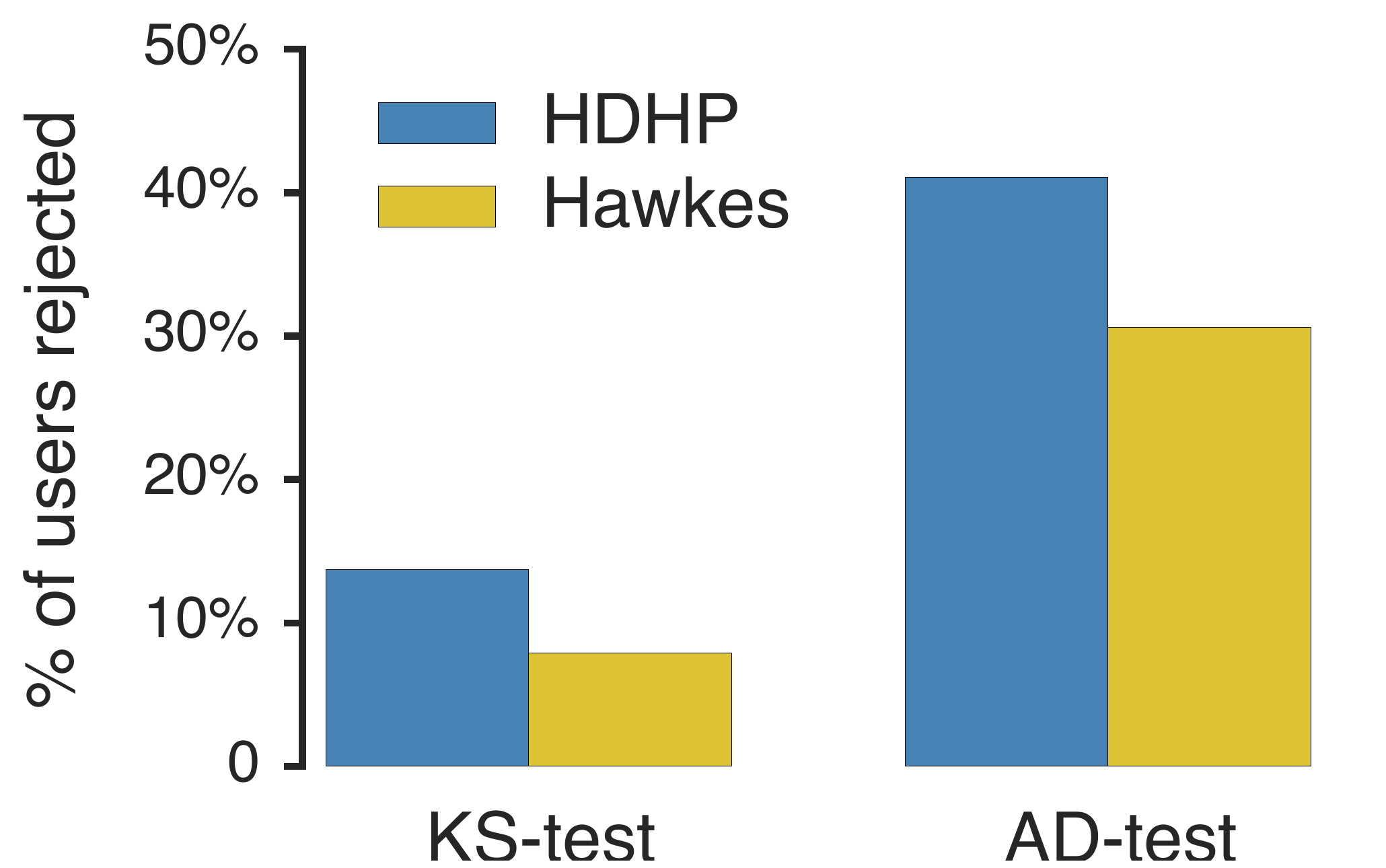}}\label{fig:test_histograms}}
	\hspace{3em}
	\subfloat[Content]
	{\makebox[0.3\linewidth][c]{\includegraphics[width=0.26\linewidth]{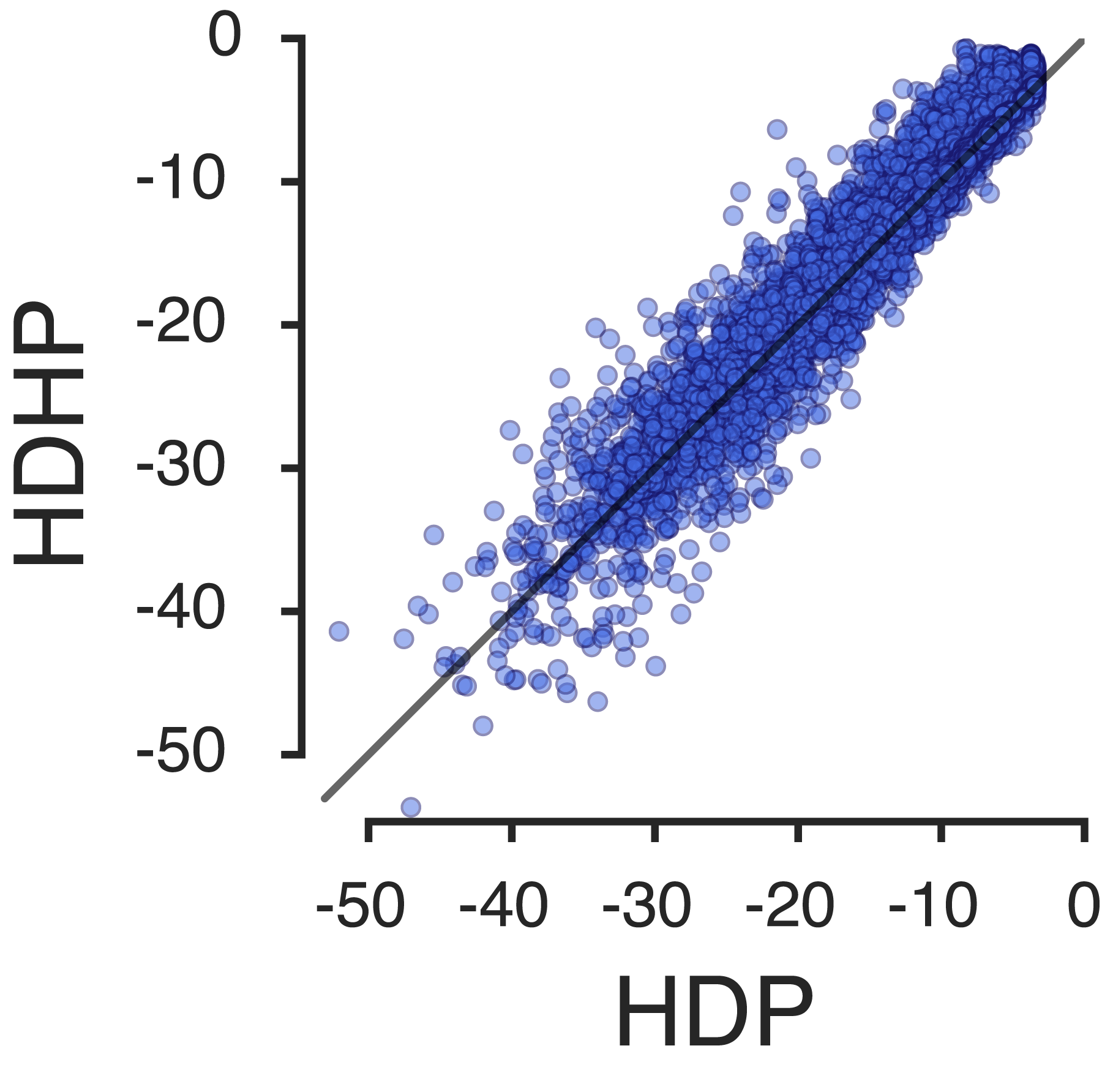}}\label{fig:full_loglike}}

	%
	\caption{Goodness of fit of the HDHP model in terms of (a) dynamics  and (b) content.}
	\label{fig:real-loglike}
	%
\end{figure}

\xhdr{Experimental setup}
We gather the times and content of all the questions 
posted by all Stack Overflow users during a four year period, from September 1, 2010 to September 1 2014. 
Here, we consider each user'{}s question as a learning action.
The reason for this choice is primarily the availability of public datasets, and, secondarily, the fact that a question provides clear evidence of the user's current interest at the time of asking. By looking only at the questions, we are underestimating the number of actions taken on each task, however, this bias is shared across all the tasks and, thus, we can still compare the dynamics of different patterns in a sensible way. For each question, we use the set of (up to) five tags (or keywords) that 
the user used to describe her question as the content associated to the learning action.
%
%
To ensure that the inferred parameters are reliable and accurate, we only consider users who posted at least 50 questions and tags that were used in at 
least 100 questions. After these preprocessing steps, our dataset consists of $\sim$$1.6$ million questions performed by $\sim$$16{,}000$ users, and a 
vocabulary of $\sim$$31{,}400$ tags. 
Finally, we run our inference algorithm on the first 45 months of data and evaluate its performance on the last three months, used as held-out set. 
In our experiments we set the time scale to be one month, the kernel decay $\nu = 5$ and the number of particles $|\Pcal|= 200$ particles, which worked well in 
practice. 
%
Our implementation of the SMC algorithms for the proposed HDHP and the HDP  requires, respectively, 71ms and 65ms per question on average, which implies that accounting for the temporal information in the data leads to an increase in runtime of approximately 10\%.  

\begin{figure*}[t]
	\centering
	\hspace{-10mm}
	\begin{tabular}{ccc}
		\subfloat{\makebox[0.3\textwidth][c]{\includegraphics[width=0.14\textwidth, trim=0cm 2cm 4cm 4cm]{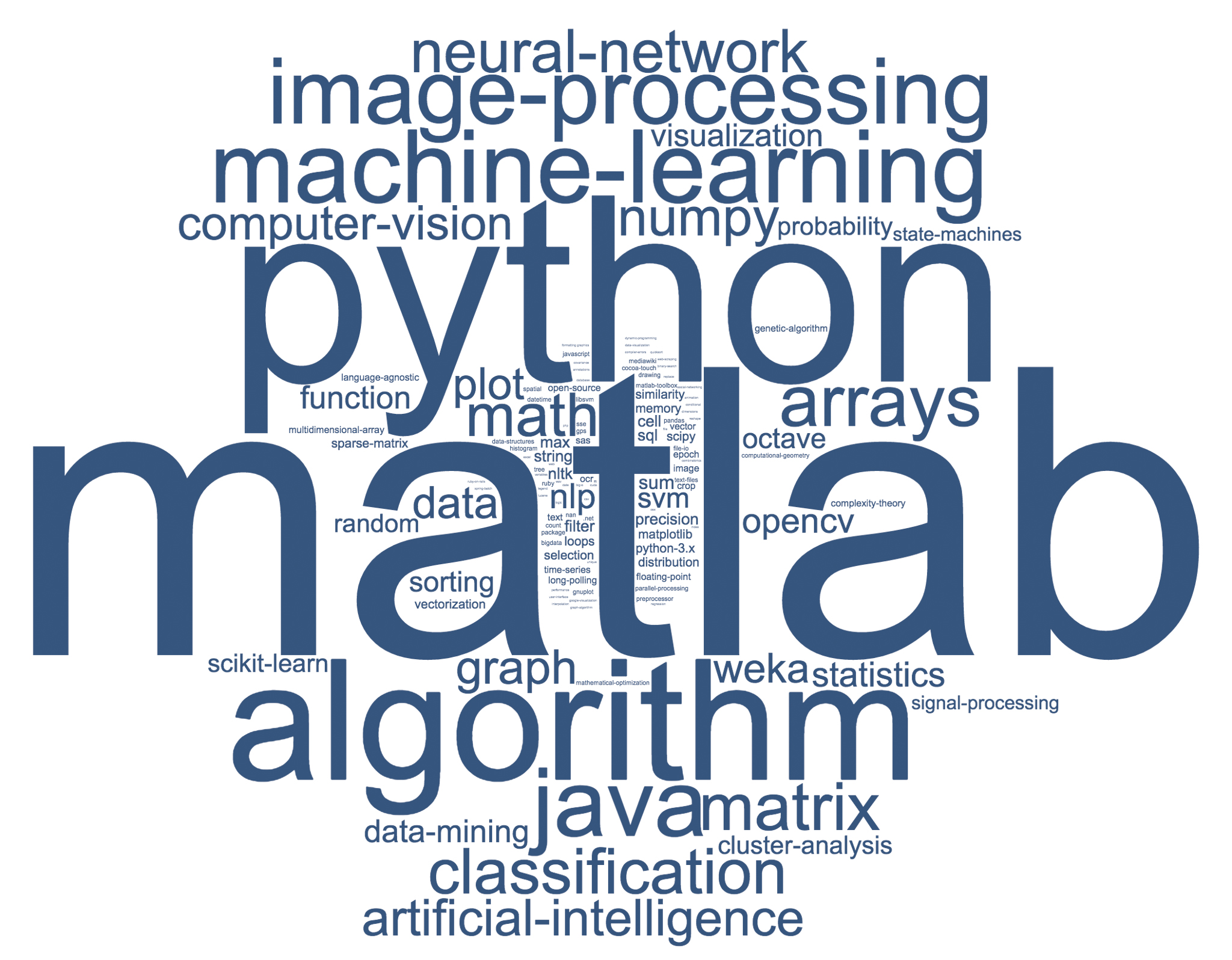}}\label{fig:machine_learning_cloud}}
		& \subfloat{\makebox[0.3\textwidth][c]{\includegraphics[width=0.16\textwidth, trim=0cm 2cm 4cm 4cm]{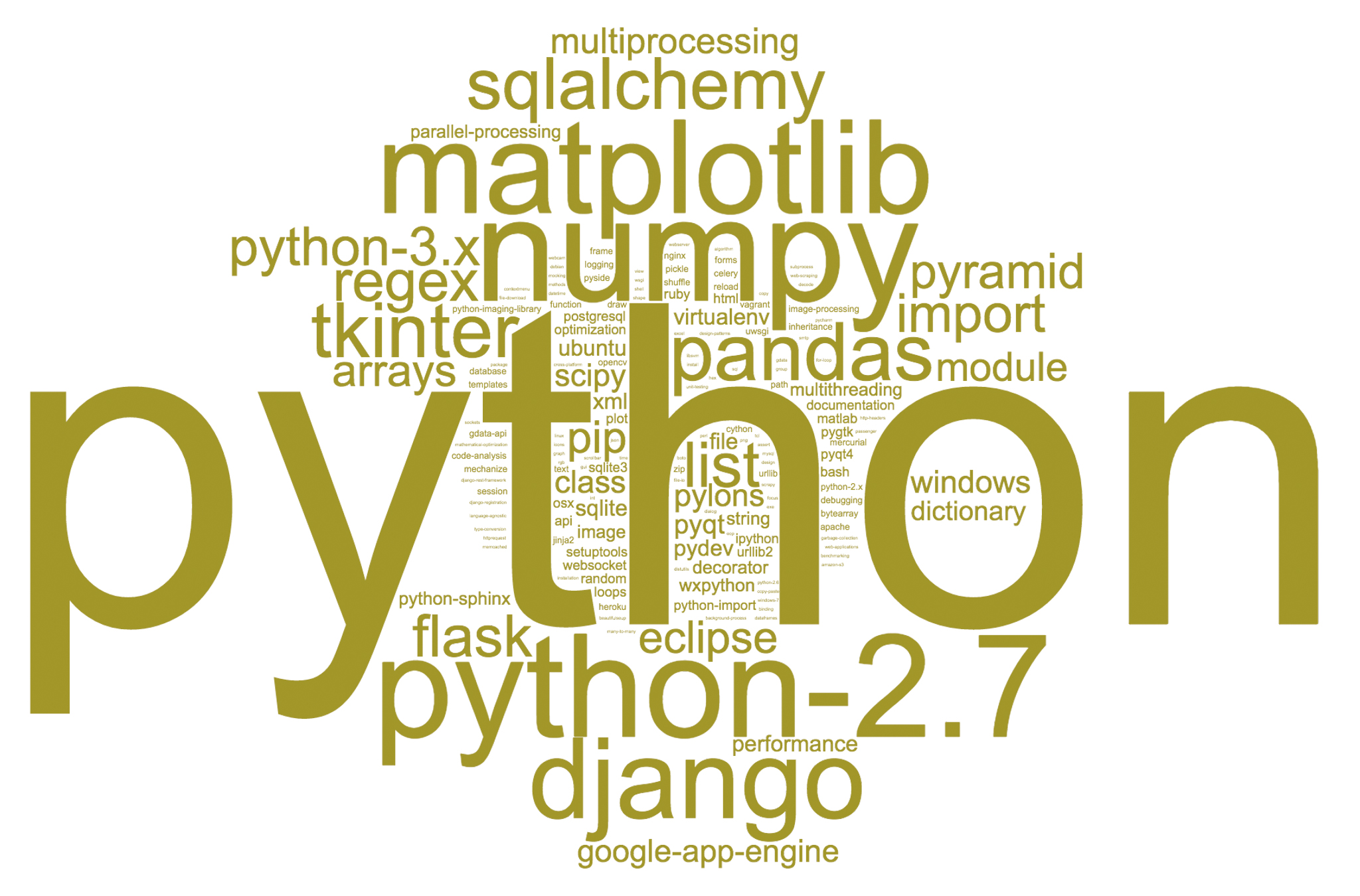}}\label{fig:python_cloud}} 
		& \subfloat{\makebox[0.3\textwidth][c]{\includegraphics[width=0.1\textwidth, trim=0cm 3cm 4cm 4cm]{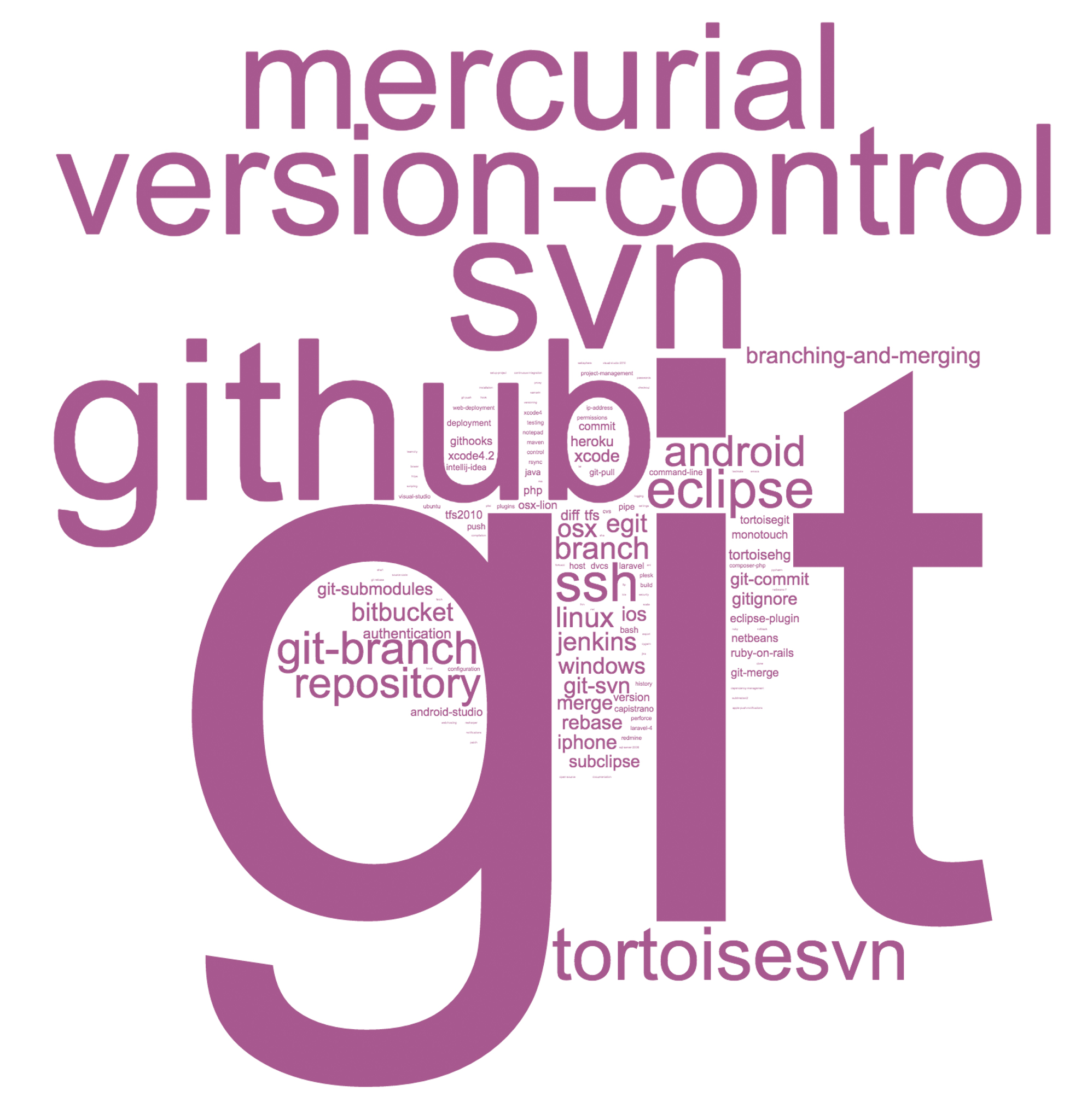}}\label{fig:version_control_cloud}} 
		\\
		
		\subfloat[Machine Learning]{\setcounter{subfigure}{1}  \makebox[0.3\textwidth][c]{\includegraphics[width=0.27\textwidth, trim = 2cm 1cm 4cm 2cm]{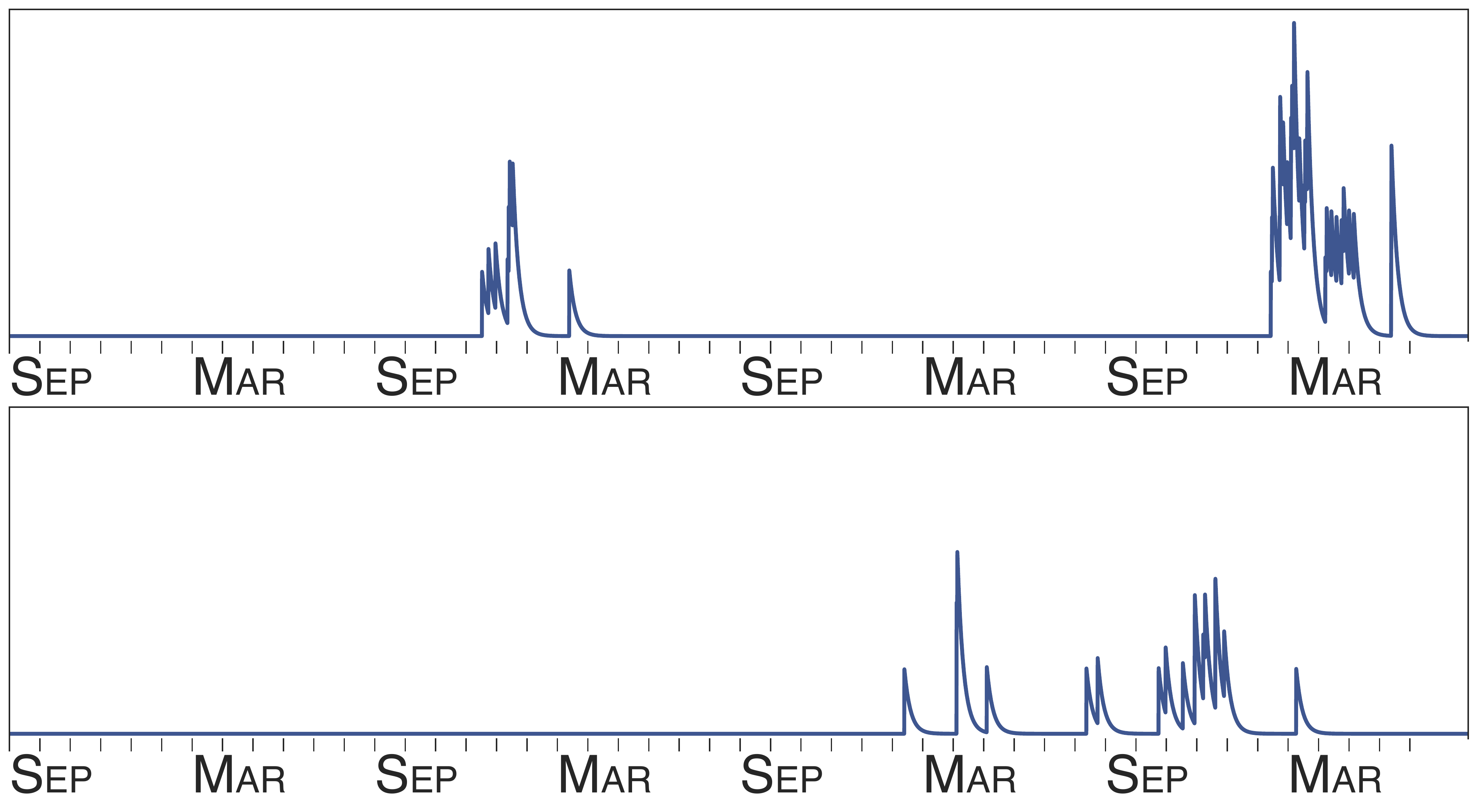}}\label{fig:machine_learning_timeline}}
		
		& \subfloat[Python]{\makebox[0.3\textwidth][c]{\includegraphics[width=0.27\textwidth, trim = 2cm 1cm 4cm 2cm]{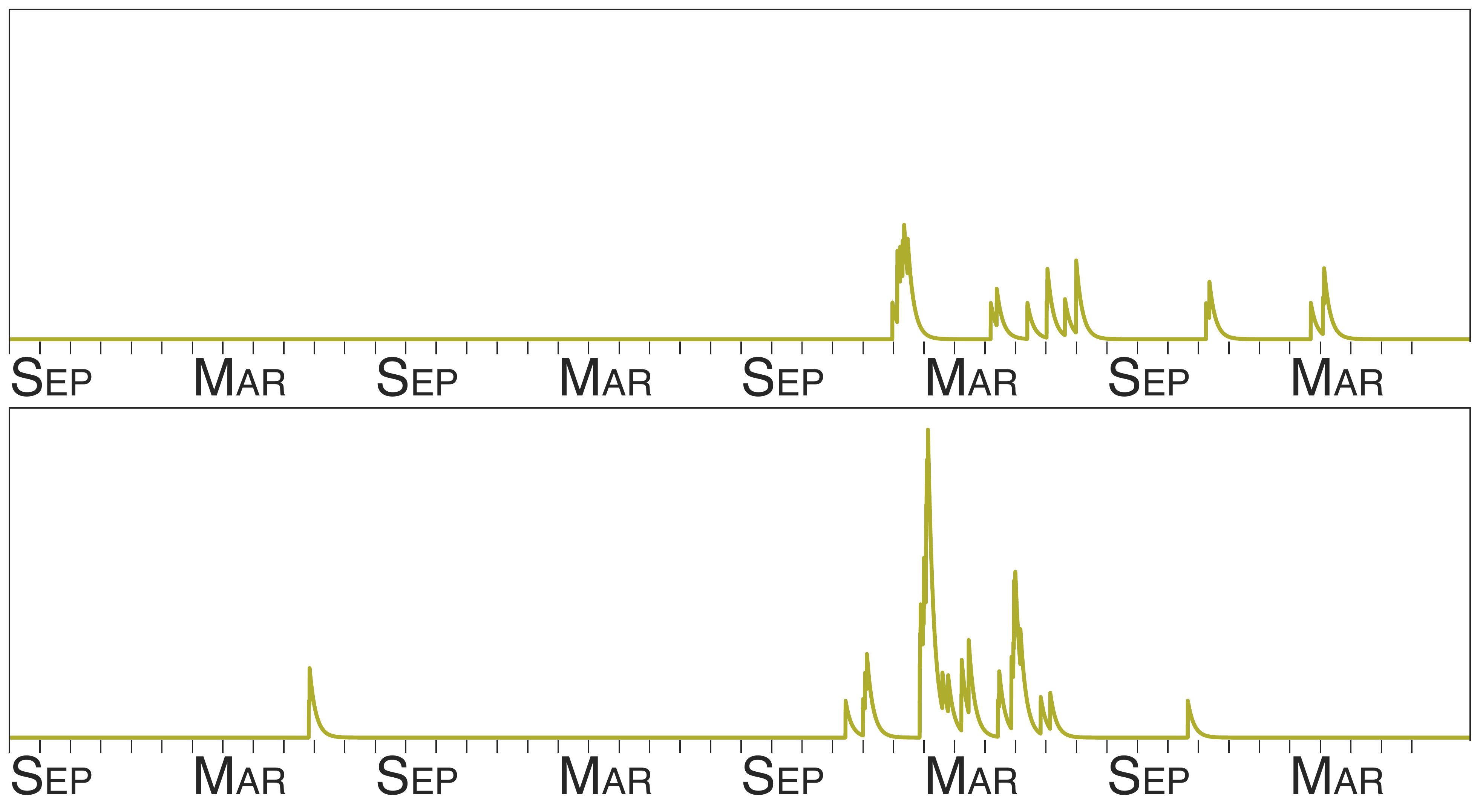}}\label{fig:python_timeline}} 
		
		&\subfloat[Version Control]{\makebox[0.3\textwidth][c]{\includegraphics[width=0.27\textwidth, trim = 2cm 1cm 4cm 2cm]{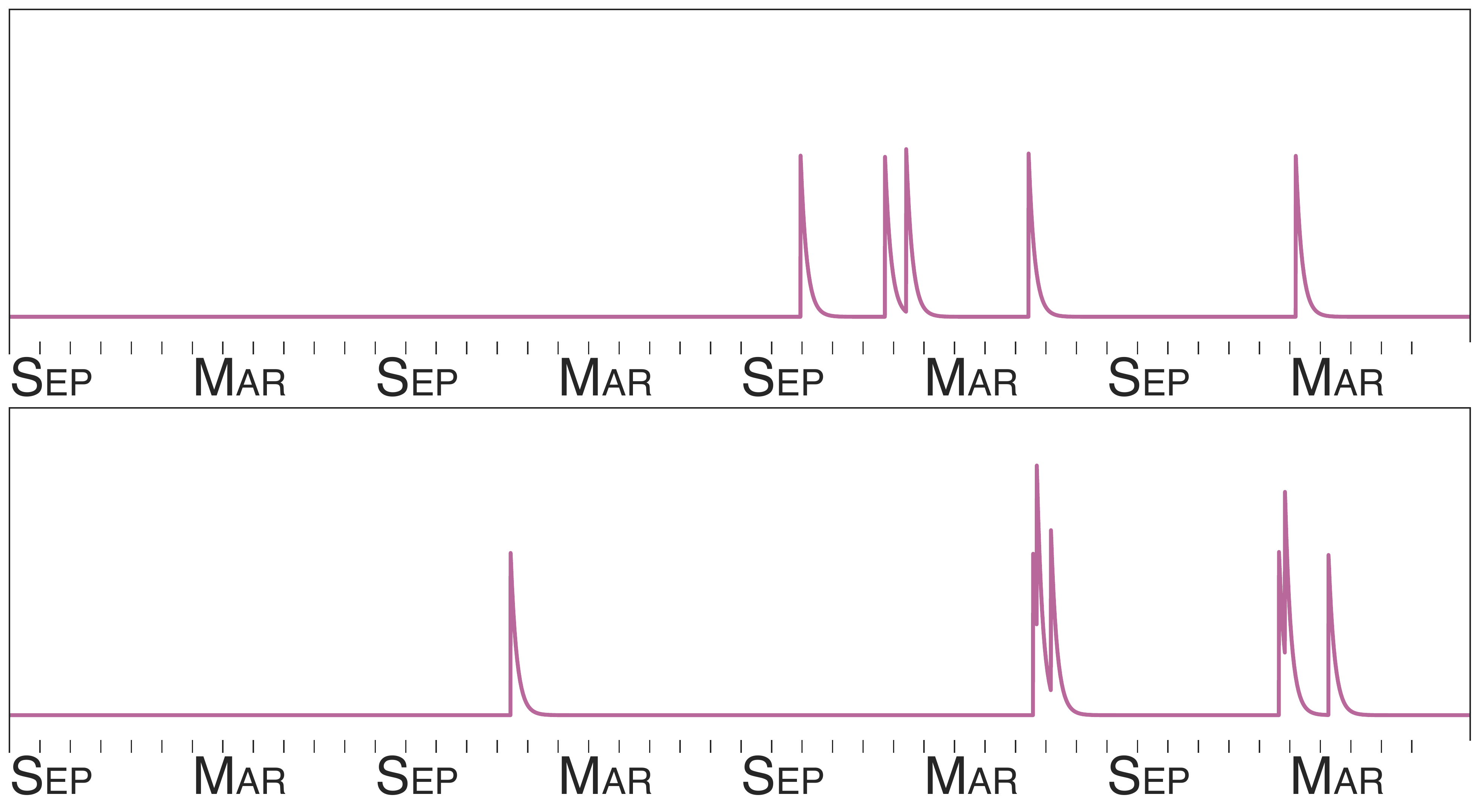}}\label{fig:version_control_timeline}}\\
	\end{tabular}
	\caption{Three inferred learning patterns in Stack Overflow. The top row shows the content associated to each pattern, in the form of clouds of words, while the bottom row shows two samples of its
		characteristic temporal dynamics, by means of the intensities of two users using the pattern.}\label{fig:topics_examples}
	%
\end{figure*}

\xhdr{Goodness of fit}
We evaluate the goodness of fit of our proposed model on learning activity data, in terms of both content and temporal dynamics. 
To this end, we first evaluate the performance of the HDHP at capturing the temporal dynamics of the learning activity and compare it with the standard Hawkes process that only accounts for the temporal information of the data and, therefore,  cannot cluster learning actions into learning patterns. 
In the latter, we model the learning activity of each user as an independent univariate Hawkes process, disregarding the content of each learning action. In other words, for each user, we learn both a base 
intensity $\mu$ and a self-excitation parameter $\alpha$, as defined in Eq.~\ref{eq:hawkes}, per user active in the test set $\Ucal_{test}$.  
In order to compare the performance of the models, we first apply the time changing theorem~\cite{daley2008pointprocessesII}, which states that the integral of the intensity of a point process between two consecutive events $\int_{t_i}^{t_{i+1}} \lambda_u^*(t) dt$ should conform to the unit-rate exponential distribution. Then, we resort to two goodness of fit tests, the Kolmogorov-Smirnov and the Anderson-Darling~\cite{stephens1978goodness}, to measure how well the transformed action times fit the target distribution. 
Figure~\ref{fig:test_histograms} summarizes the results by showing the percentage of the users in the held-out set that each test rejects at a significance level of $5\%$. While the Hawkes process performs 
slightly better (5\% for the KS-test and 11\% for the AD-test) than our model, it does so by using almost $2\times$ more parameters --- $2|\Ucal_{test}| \sim 5$k for the Hawkes  vs $|\Ucal_{test}| + L^* \sim 2.7$k for
the HDHP, where  $L^*=227$ is the number of inferred 
learning patterns. 
%

Second, we focus on evaluating the performance of the HDHP at clustering learning activity, and compare it with the HDP~\cite{Teh2006}, which only makes use of the content information in the data.  
We resort to the marginal likelihood of the inferred parameters evaluated on the held-out set of questions $\qb_i \in \Dcal_{test}$, \ie, 
\begin{equation}
p(\qb_i |  \Dcal_{train}, u_i, t_i)= \sum_{\ell=1}^{L}  p(\qb_i | \Dcal_{train}, \ell) p(\ell | \Dcal_{train}, u_i, t_i).\nonumber 
\end{equation} 
Above, the first term is defined in the same way for both models. However, for the HDP, the second term is simply the topic popularity $\pi_\ell$, while for the HDHP it 
depends on the complete user history up to but not including $t_i$, \ie, $p(\ell | \Dcal_{train}, u_i, t_i) \propto {\lambda^*_{u_i,\ell}(t_i)} $,
where $\lambda^*_{u_i,\ell}(t_i)$ is given by Eq.~\ref{eq:pattern_intensity}. Figure~\ref{fig:full_loglike} shows the log-likelihood values obtained under the proposed 
HDHP and the HDP on the held-out set. Here, higher log-likelihood values mean better goodness of fit and, therefore, all the points above the $x=y$ line correspond to 
questions that are better captured by the HDHP, which are in turn 60\% of the held-out questions.
Additionally, we also compute the perplexity~\cite{LDA} as 
\begin{equation}
perplexity= \exp \left\{ - \frac{ \sum_{i:e_i\in \Dcal_{test}} \log p(\qb_i | \Dcal_{train})}{ | \Dcal_{test}|} \right\}. \nonumber 
\end{equation}
%
The perplexity values for the HDHP and the HDP are $204$ and $243$, respectively, where here lower perplexity values mean better goodness of fit. These results show that by modeling temporal 
information, in addition to content information, the HDHP fits better the content in the data than the HDP (20\% gain in perplexity), and therefore, it provides more meaningful learning patterns in terms of content.
%

\xhdr{Learning patterns}
In this section, our goal is to understand the characteristic properties of the learning patterns that Stack Overflow users follow for problem solving. 
To this end, we first pay attention to three particular examples of learning patterns, `\emph{Machine Learning}', `\emph{Python}' and `\emph{Version Control}', among the uncovered $L^*=227$ 
learning patterns, and investigate their characteristic properties. 
Figure~\ref{fig:topics_examples} compares the above mentioned patterns in terms of content, by means of word clouds, and in terms of temporal dynamics, by means of the learning 
pattern intensities associated to two different users active on each of the patterns. 
Here, we observe that: i) the cloud of words associated to each inferred learning pattern corresponds to meaningful topics; and ii) despite the stochastic nature of the temporal dynamics, 
the user intensities within the same learning pattern tend to exhibit striking similarities in terms of burstiness and periods of inactivity. 
For example, we observe that the \emph{Machine Learning} and \emph{Python} tasks exhibit much larger bursts of events than \emph{Version Control}. 
A plausible explanation is that version control problems tend to be more specific and simple, \eg, resolving a conflict while merging versions, and, thus, can  be quickly solved with one or just a few questions. On the contrary, a user interested in machine learning or Python may face more complex problems whose solution requires asking several questions in a relatively short period of time. 

\begin{figure*}[t]
\centering
\subfloat[Popularity]{\makebox[0.3\textwidth][c]{\includegraphics[width=0.3\textwidth]{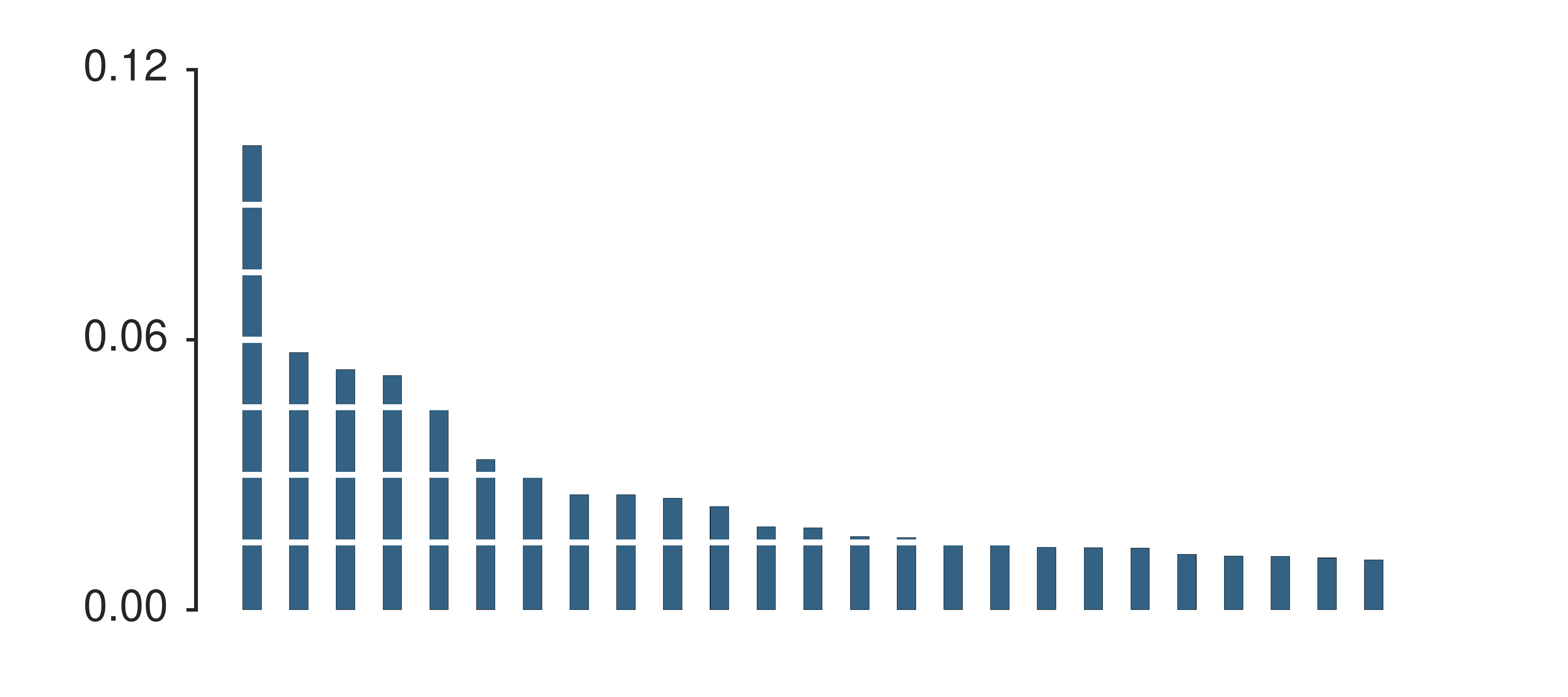}}\label{fig:popularity}}
\subfloat[Burstiness]{\makebox[0.3\textwidth][c]{\includegraphics[width=0.3\textwidth]{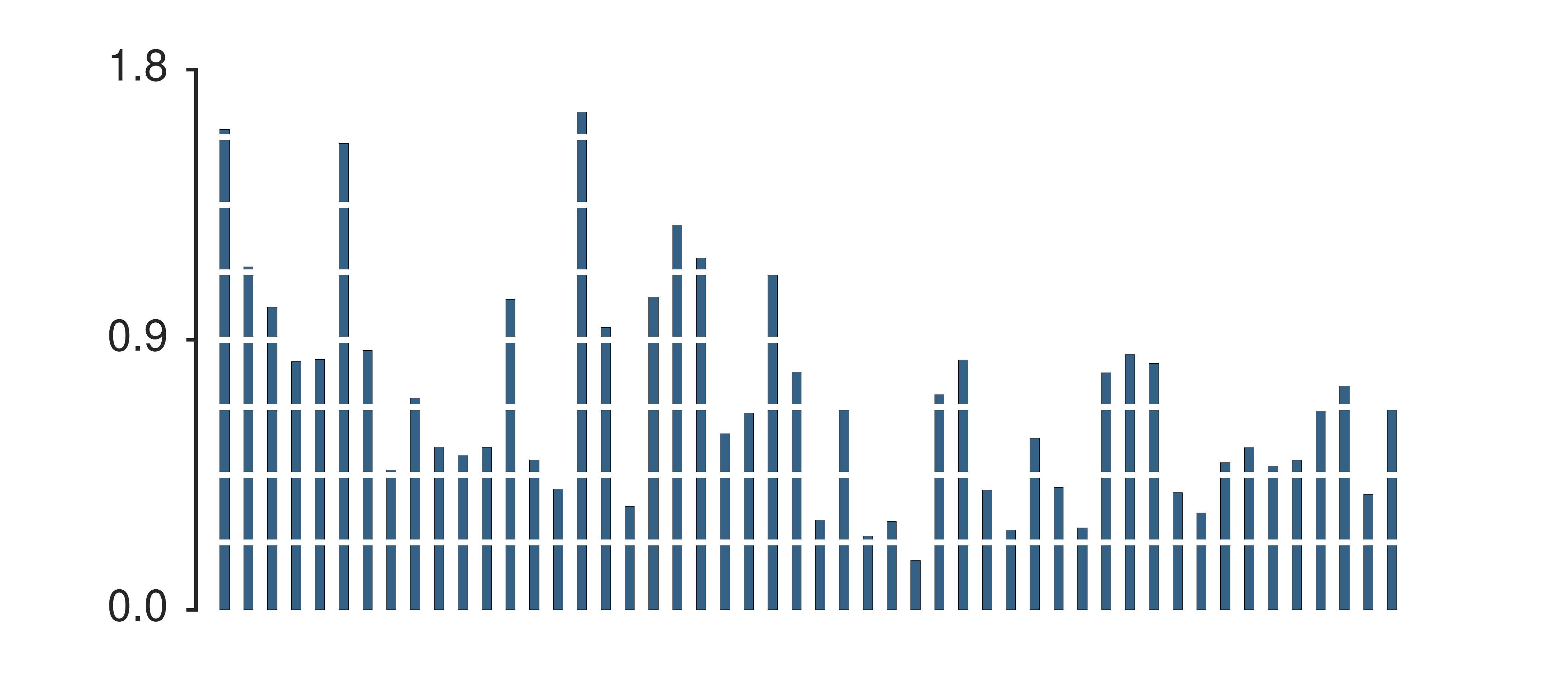}}\label{fig:burstiness}}
\subfloat[Popularity vs. Burstiness]{\makebox[0.3\textwidth][c]{\includegraphics[width=0.3\textwidth]{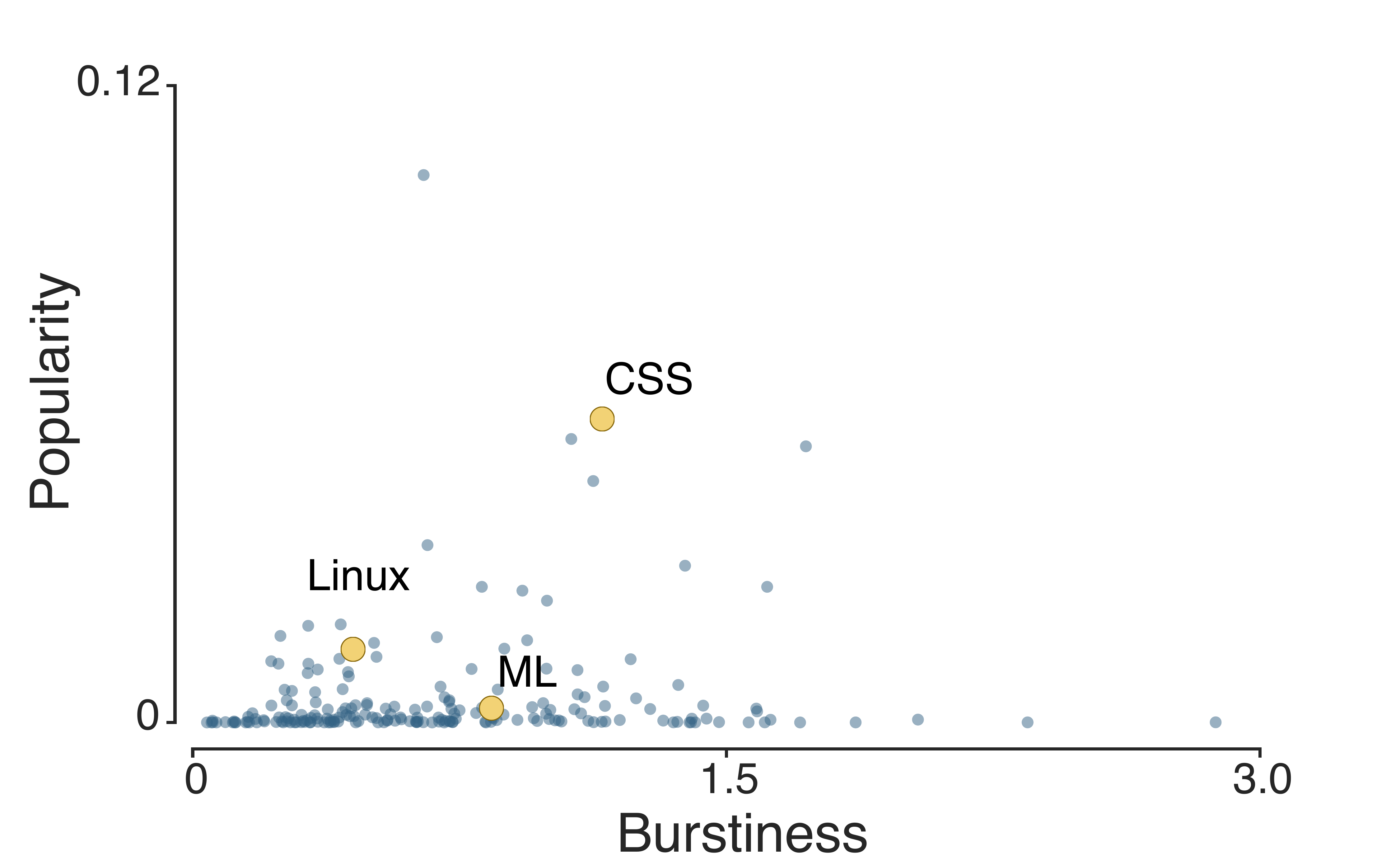}}\label{fig:popularity_vs_burstiness}}
%
\caption{Learning patterns. Panels (a) and (b) show the popularity and burstiness of the top-50 most popular learning patterns, and panel (c) shows the popularity and burstiness for all the inferred patterns. We highlight the learning pattern examples in Figure~\ref{fig:topics_examples}, as well as some others from Table~\ref{tbl:patterns}.}\label{fig:pop_bursty}
%
\end{figure*}

Next, we investigate whether more popular learning patterns are also the ones that trigger larger bursts of events, \ie, the learning patterns that engage users to 
perform long sequences of closely related learning actions in shorts period of time. 
Figure~\ref{fig:pop_bursty} shows the popularity and burstiness of the 50 most popular learning patterns sorted in decreasing order of popularity, as well as a scatter plot which shows the popularity against burstiness for all the inferred patterns.
Here, we compute the burstiness as the expected number of learning events triggered by self-excitation during the first month after the adoption of the pattern using Eq.~\ref{eq:N_u}. 
Figure~\ref{fig:pop_bursty} reveals that the burstiness is not correlated with the popularity of each pattern. On the contrary, even among the top 20 most popular patterns, several learning patterns trigger on average less than 0.5 follow-up questions. It is also worth noticing that there is a small set of learning patterns which are much more popular than the rest. 
In particular, the most popular learning pattern, which is related to 
\emph{Web design}, captures approximatively 12\% of the attention of Stack Overflow users, and the 20 most popular learning patterns gather more than 60\% of the popularity.
Moreover, Figure~\ref{fig:popularity_vs_burstiness} 
highlights examples of learning patterns that are very popular and 
bursty -- \emph{Web design}; examples of bursty learning patterns that are not very popular -- \emph{machine learning}; and learning patterns that are not popular nor bursty -- \emph{UI libs}. 
%
Table~\ref{tbl:patterns} shows the top-20 most probable words in the seven learning patterns highlighted in Figure~\ref{fig:popularity_vs_burstiness}.

\xhdr{User behavior}
In this section, we use our model to identify different types of users and derive insights about the learning patterns they use over time, as well as the evolution of their interests. Two natural questions emerge in this context: (i) Do users stick to just a few learning patterns for all their tasks, or perhaps they explore a different pattern every time they start a task? And, (ii) how long do they commit on their chosen task? 

\begin{figure*}[t]
	\centering
	\subfloat[Base intensity $\mu_u$ (tasks/month)]
	{\makebox[0.33\textwidth][c]{\includegraphics[width=0.25\textwidth, trim=2cm 0mm 0 0]{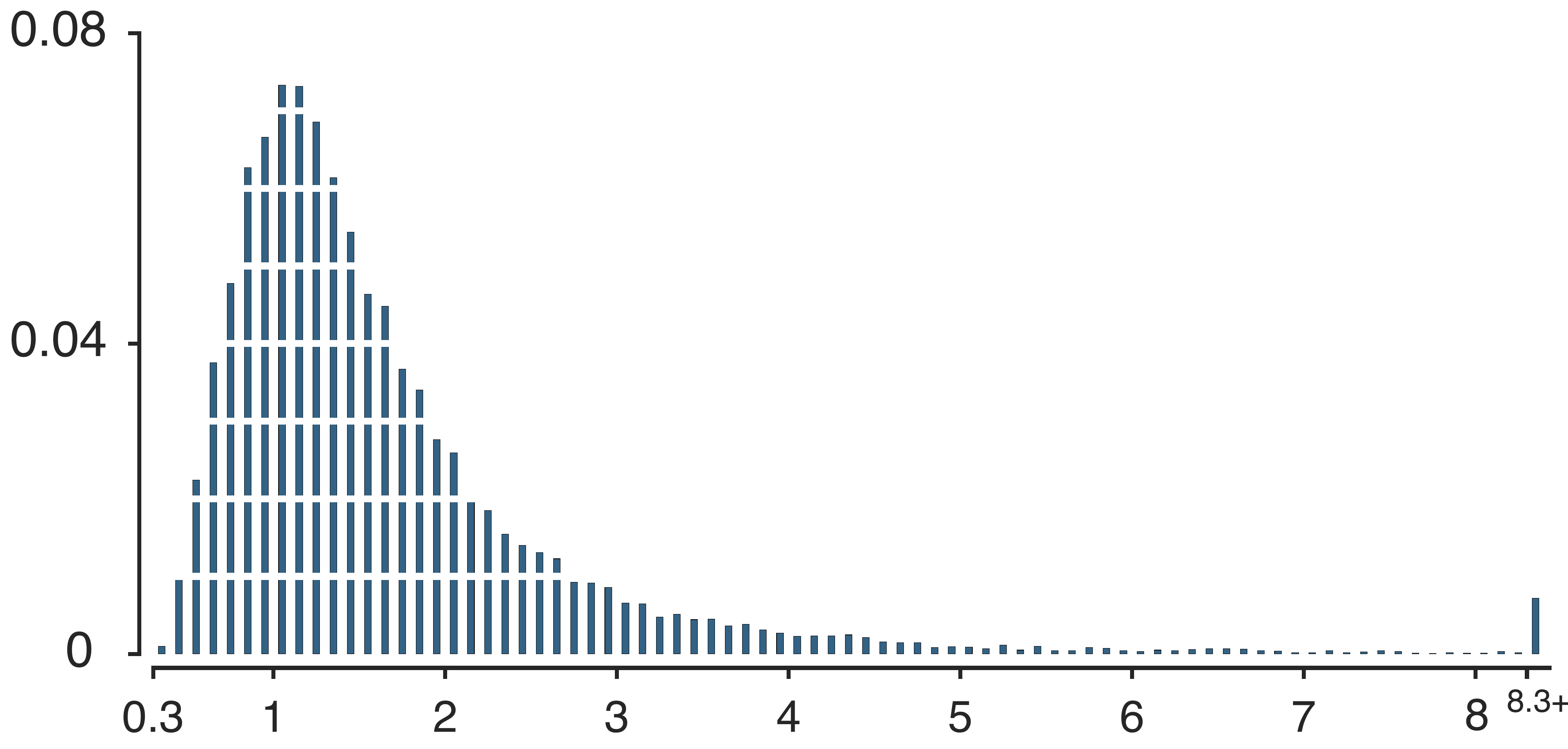}\vspace{30mm}}\label{fig:user_mu}}
	\subfloat[\# of adopted patterns per user]
	{\makebox[0.33\textwidth][c]{\includegraphics[width=0.25\textwidth, trim=2cm 0mm 0 0]{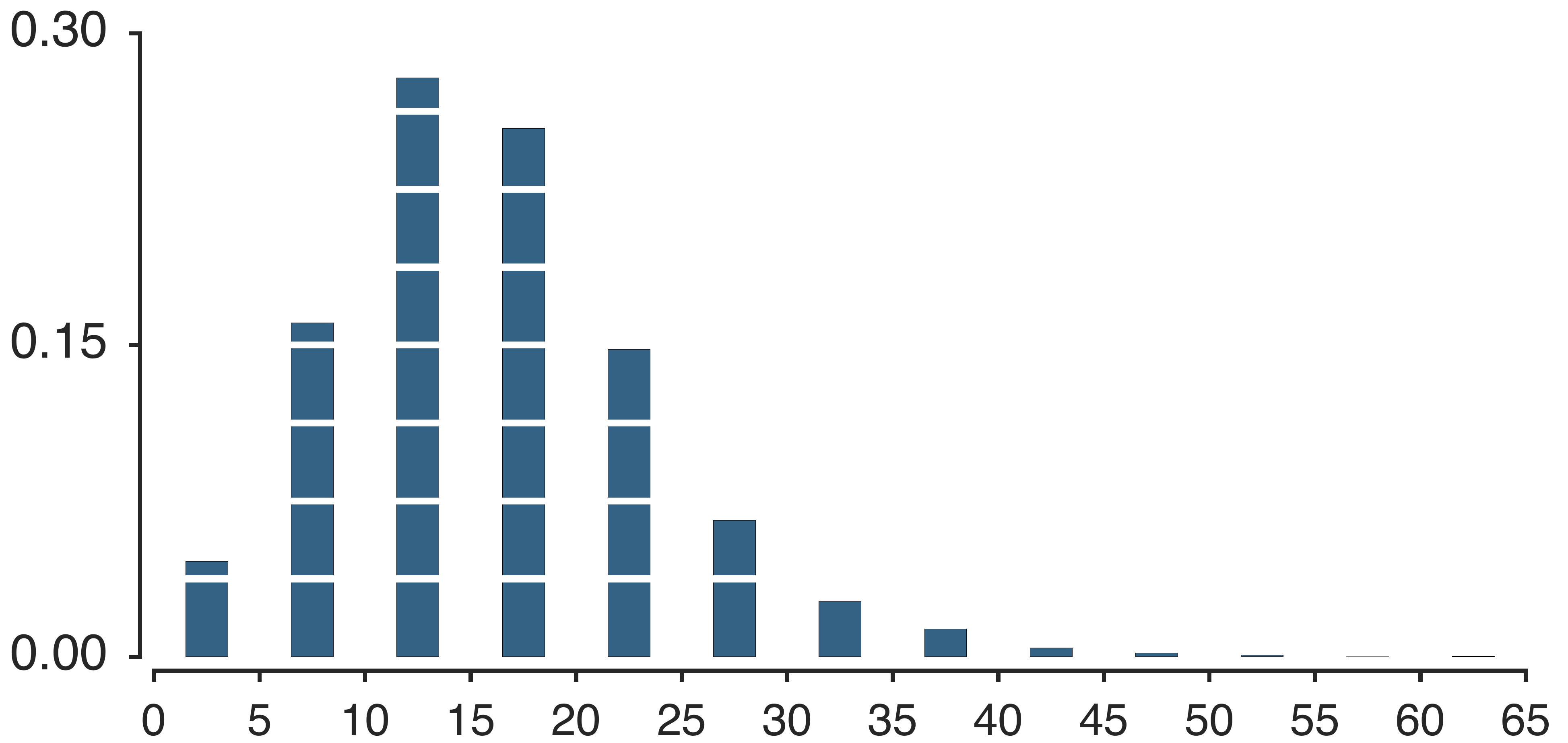}}\label{fig:patterns_per_user}}	
	\subfloat[Average time in months per task]{\makebox[0.33\textwidth][c]{\includegraphics[width=0.28\textwidth]{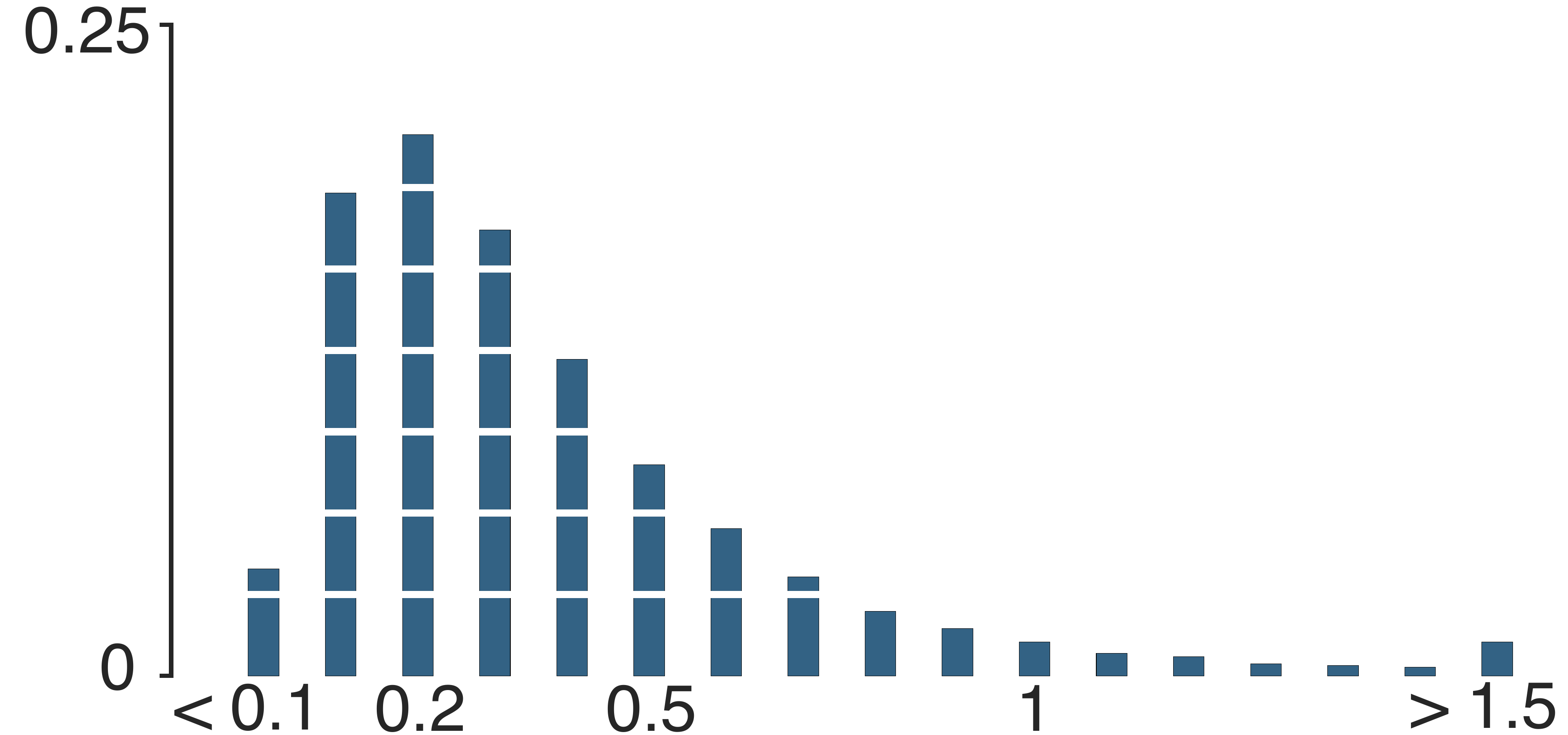}}\label{fig:average_time_on_pattern}}
	\caption{User behavior: (a) the inferred user base intensities, (b) the number of learning patterns adopted by the users over the 4 years, and (c) the average time users spent for the completion of their tasks.}
	\label{fig:stackoverflow_analysis}
	%
\end{figure*}

\begin{figure}
	\centering
	\subfloat[Explorer user]{\makebox[0.9\linewidth][c]{\includegraphics[width=0.9\linewidth]{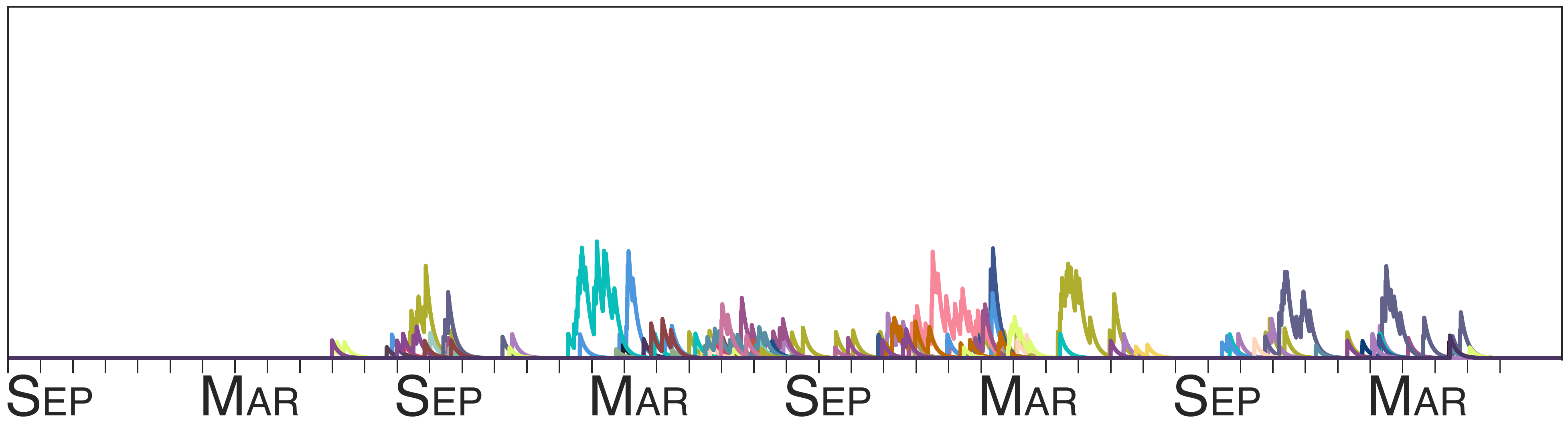}}\label{fig:explorer_user}}
	\\
	\subfloat[Loyal user]{\setcounter{subfigure}{2} \makebox[0.9\linewidth][c]{\includegraphics[width=0.9\linewidth]{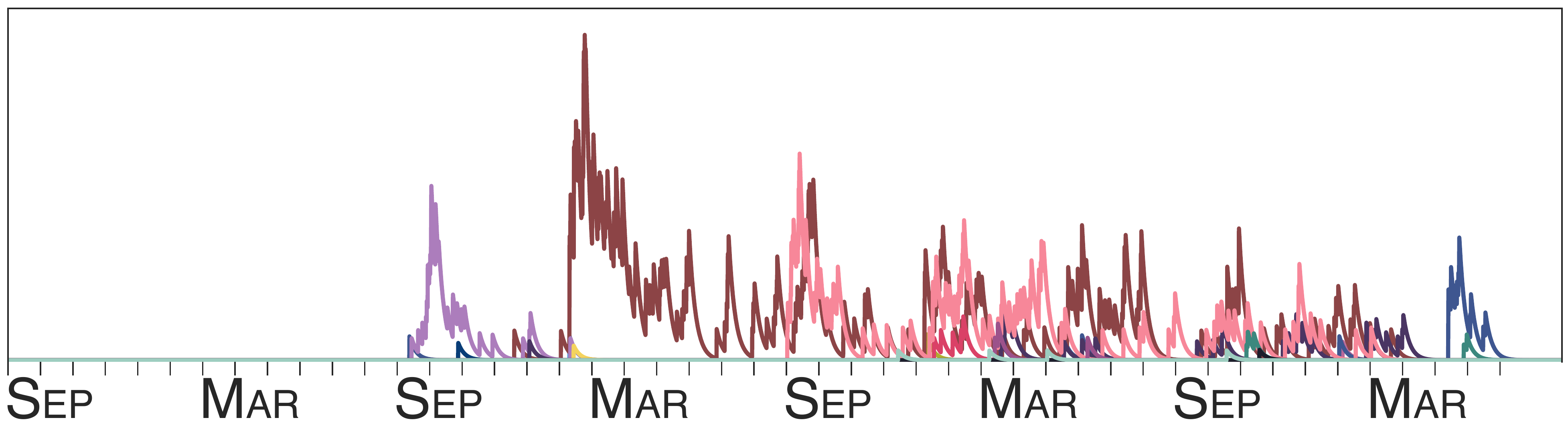}}\label{fig:loyal_user}}
	%
	\caption{Real-world examples of user behavior. An explorer user (panel (a)), shifts over many different learning patterns over time, while a loyal user (panel (b)) sticks to a small selection of patterns.}\label{fig:tracking}
	%
\end{figure}

First, we visualize the inferred intensities for two specific real users, among the several that we found, in Figures~\ref{fig:explorer_user} - \ref{fig:loyal_user}. These are examples of two very distinctive behaviors:
\begin{compactitem}
\item[--] {\bf Explorers}: They shift over many different learning patterns and rarely adopt the same pattern more than once. For example, the
user in Figure~\ref{fig:explorer_user} adopts over 10 patterns in less than a year, and rarely adopt the same learning pattern more than once. 

\item[--] {\bf Loyals}: They remain loyal to a few learning patterns over the whole observation period. For example, the user in Figure~\ref{fig:loyal_user} asks questions associated to two learning patterns over a period of 4 years period and rarely adopts new learning patterns.
\end{compactitem}
We investigate to which extent we find explorers and loyals throughout Stack Overflow at large. To this end, we compute the user base intensities, $\mu_u$, which can be viewed as the number of new tasks that a 
user starts per month, and the distribution of the total number of learning patterns adopted by each user over the observation period.
Figures~\ref{fig:user_mu} and \ref{fig:patterns_per_user} summarize the results, showing several interesting patterns. First, there is a high variability across users in terms of new task rate -- while most of users start one to two new tasks every month, there are users who start up to more than 8 tasks monthly. Second, while approximately 5\% of the users remain loyal to at most 5 learning patterns and another 10\% of the user explores more than 25 learning 
patterns over the 4 years, the average user ($\sim$87\%) adopts between 5 and 25 patterns.   

Finally, we investigate how long do users commit on their chosen task. To answer this question, we compute the average time between the initial and the final event for each task of our users. Figure~\ref{fig:average_time_on_pattern} show the distribution of the average time spent per task.
Here we observe that while approximatively 10\% of the user tasks are concluded in less than a month, most of the users (over 75\% of the users) spend one to four months to complete a task.    

\section{Conclusions}
%
%
In this paper, we proposed a novel probabilistic model, the Hierarchical Dirichlet Hawkes Process (HDHP),  for clustering grouped streaming data. In our application, each group corresponds to a specific user{}'s 
learning activity. The clusters correspond to learning patterns, characterized by both the content and temporal information and shared across all users. We then developed an efficient  
inference algorithm, which scales linearly with the number of users and learning actions, and  accurately recovers both the pattern associated with each learning user action and the model parameters. 
Our experiments on large-scale data from Stack Overflow show that the HDHP recovers meaningful learning patterns, both in terms of content and temporal dynamics, and offers a characterization of different user behaviors.
 
%
%
We remark that, the proposed HDHP could be run within the learning platform in an online fashion to track users' interest in real time. With this, one could get both a characterization of the different user behaviors, and recommendations on questions that might be of interest at any given time. 
Finally,  although here we focused on modeling online learning activity data, the proposed HDHP can be easily used to cluster a wide variety of streaming data, ranging from news articles, in which there is a single stream (group) of data, to web browsing, where one could identify groups of websites that provide similar services or content. 

{ \small
\bibliographystyle{abbrv}
\bibliography{Bib} }

\newpage
\appendix
\begin{table}[t]
	\renewcommand{\arraystretch}{1.2} 
	\centering
	\begin{tabular}{p{\linewidth}} 
		\toprule \small
		Top-$20$ most probable words in the learning pattern\\
		\midrule
		\small	\textit{\textbf{`Web design':}} jquery javascript html php css ajax jquery-ui json forms arrays asp.net html5 jquery-mobile mysql dom regex jquery-plugins internet-explorer jquery-selectors wordpress \vspace{1mm}\\
		
		\small	\textit{\textbf{`sql':}} sql mysql sql-server php sql-server-2008 database tsql oracle postgresql sql-server-2005 database-design join stored-procedures c\# select sqlite sql-server-2008-r2 java performance datetime \vspace{1mm}\\
		
		\small	\textit{\textbf{`iOS':}} ios objective-c iphone xcode cocoa-touch ipad uitableview cocoa core-data osx ios4 ios5 uiview uitableviewcell uiviewcontroller ios7 ios6 uinavigationcontroller uiscrollview nsstring \vspace{1mm}\\
		
		\small	\textit{\textbf{`Python':}} python numpy python-2.7 matplotlib django pandas python-3.x scipy tkinter flask sqlalchemy list arrays wxpython regex dictionary multithreading osx import google-app-engine \vspace{1mm}\\
		
		\small	\textit{\textbf{`Version control':}} git svn github version-control mercurial eclipse tortoisesvn merge branch repository ssh bitbucket xcode git-branch commit git-svn osx windows java gitignore\vspace{1mm}\\
		
		\small	\textit{\textbf{`Machine learning' (ML):}} matlab python algorithm r machine-learning java matrix plot artificial-intelligence numpy arrays image-processing nlp statistics opencv math octave data-mining scikit-learn neural-network\vspace{1mm}\\
		
		\small	\textit{\textbf{`UI Libraries':}} knockout.js javascript kendo-ui jquery asp.net-mvc knockout-2.0 kendo-grid durandal asp.net-mvc-4 knockout-mapping-plugin kendo-asp.net-mvc breeze single-page-application typescript mvvm asp.net-mvc-3 data-binding signalr json twitter-bootstrap \\
		
%
%
%
		\bottomrule
	\end{tabular}
	\caption{The $20$ most probable words for the seven patterns highlighted in Figure~\ref{fig:popularity_vs_burstiness}.}\label{tbl:patterns}
	  \vspace{-3mm}
\end{table}

\begin{algorithm}[t]
\small
  Initialize $w^{(p)}_1= 1/ |\Pcal |$, $K^{(p)}=0$ and $L^{(p)}=0$ for all $p \in \Pcal$. \\
  \For{$i=1,\ldots,n$}{
  \For{$p \in \Pcal$}{
%
    Update the kernel parameters $\alpha^{(p)}_\ell$ for $\ell=1,\ldots,L^{(p)}$ and the user base intensities $\mu^{(p)}_u$ for all $u \in \Ucal$.\\
    Draw $b^{(p)}_i$ from~\eqref{eq:sample-eq}.

    \If {$b^{(p)}_i =K^{(p)}+1$}{
    	 Draw the new task parameters $\phi^{(p)}_{K^{(p)}+1}$ according to~\eqref{eq:phi_sample}. \\
	 Increase the number of tasks $K^{(p)}= K^{(p)}+1$.\\
	 \If {$\phi^{(p)}_{K^{(p)}+1} = \varphi^{(p)}_{L^{(p)}+1}$}{
	 Draw the triggering kernel for the new topic $\alpha^{(p)}_{L^{(p)}+1}$ from the prior. \\
	 Increase the total number of learning patterns $L^{(p)}= L^{(p)}+1$.
	 }
    }
    Update the particle weight $\textrm{w}^{(p)}_i$ according to \eqref{eq:wn}.
   }
    Normalize particle weights. \\
     \If {$\|\textrm{\bf w}_i\|_2^2<threshold$}{
     Resample particles.

     }
 }
  \caption{Inference algorithm for the HDHP\label{alg:inference}}
\end{algorithm}

\def\thesection{\Alph{section}}
\section*{Appendix}
\section{Details on the Inference} \label{app:inference} 
Given a collection of $n$ observed learning actions performed by the users of an online learning site during a time period $[0, T)$, our
goal is to infer the learning patterns that these events belong to. 
To efficiently sample from the posterior distribution, we derive a sequential Monte Carlo (SMC) algorithm that exploits the temporal dependencies in the observed
data to sequentially sample the latent variables \-associated with each learning action.
In particular, the posterior distribution $p(b_{1:n} |t_{1:n}, u_{1:n}, q_{1:n})$ is sequentially approxi\-ma\-ted with a set of $|\Pcal |$ particles, which are sampled from a
proposal distribution that factorizes as
\begin{equation}
q(b_{1:n} |  t_{1:n},  u_{1:n}, \qb_{1:n}) =  q(b_n|b_{1:n-1} , t_{1:n},  u_{1:n}, \qb_{1:n})  q(b_{1:n-1} | t_{1:n-1},  u_{1:n-1}, \qb_{1:n-1}),\nonumber
\end{equation}
where
\begin{equation}
q(b_{n} | \qb_{1:n}, t_{1:n}, u_{1:n})  =\frac{p( \qb_n | \qb_{1:n-1},  b_{1:n})  p(b_{n} |  t_{1:n}, u_{1:n}, b_{1:n-1})}{\sum_{(b_n)}p( \qb_n | \qb_{1:n-1},  b_{1:n})  p(b_{n} |  t_{1:n}, u_{1:n}, b_{1:n-1})} 
\label{eq:sample-eq}
 \end{equation}
 In the above expression, we can exploit the conjugacy between the multinomial and the Dirichlet distributions to integrate out the word distributions $\thetab_\ell$ and
 obtain the marginal likelihood
 \begin{equation}
 p( \qb_n | \qb_{1:n-1},  b_{1:n})  = \frac{\Gamma \left( C_{\ell}^{n-1} + \eta_0 |\Wcal | \right) \prod_{w \in \Wcal} \Gamma \left( C_{w, \ell}^{n} + \eta_0 \right)}
{ \Gamma \left( C_{\ell}^{n} + \eta_0  |\Wcal |  \right) \prod_{w \in \Wcal} \Gamma \left( C_{w, \ell}^{n-1} + \eta_0 \right)}, \nonumber
 \end{equation}
 where $C_{\ell}^{n-1}$ and $C_{\ell}^{n}$ are the number of words in the pieces of content (or queries) $\qb_{1:n-1}$ and $\qb_{1:n}$, respectively, and $C_{w, \ell}^{n-1}$ and $C_{w, \ell}^{n}$ count the number of times that the $w$ appears in queries $\qb_{1:n-1}$ and $\qb_{1:n}$, respectively.
For each particle $p$, the importance weight can be iteratively computed as
\begin{multline}
\label{eq:wn} 
\textrm{w}^{(p)}_n =\textrm{w}^{(p)}_{n-1}  p( t_n, u_n | t_{1:n-1}, u_{1:n-1}, b^{(p)}_{1:n-1}, \{\alpha^{(p)}_\ell \}_{\ell=1}^L) Q^{(p)}_n,\\ \hspace{-1cm} \vspace{-2mm}
\end{multline}
where $\textrm{w}^{(p)}_1= 1/ |\Pcal |$ and  
\begin{align}
Q^{(p)}_n& = \sum_{(b_n)}p( \qb_n | \qb_{1:n-1},  b_{1:n})  p(b_{n} |  t_{1:n}, u_{1:n}, b^{(p)}_{1:n-1}) \vspace{-2mm}
\end{align}
%
Since the likelihood term depends on the user base intensities $\mu_u$ and the kernel parameters $\{\alpha^{(p)}_\ell \}_{\ell=1}^L$, following the literature in SMC devoted
to the estimation of a static parameter~\cite{cappe2007overview, carvalho2010particle}, we infer these parameters in an online manner.
%
In particular, we sample the kernel parameters from their posterior  distribution up to, but not including, time $t$, 
%
and we update the user base intensities at time $t$ as $\mu^{new}_u = r \mu^{old}_u + (1-r) \hat{\mu}_u$, where $\hat{\mu}_u$ is the maximum likelihood estimation of this parameter given the user history $\Hcal_u(t)$ and $r \in [0,1]$ is a factor that controls how much the updated parameter $\mu^{new}_u$ differs from its previous value $\mu^{old}_u$.

Algorithm~\ref{alg:inference} summarizes the overall inference procedure, which presents complexity $O(\Pcal  (\Ucal + L + K)+ \Pcal )$ per learning action $i$ fed to the algorithm, where $L$ and $K$ are the total number of learning patterns and tasks inferred up to the $(i-1)$-th action. Note also that, the for-loop across the particles $p\in \Pcal$ can be parallelized, reducing the complexity per learning action to $O( \Ucal + L + K+ \Pcal )$ . 


%
%

\end{document}